\documentclass[a4paper,11pt]{article}
\usepackage{jinstpub}
\usepackage{lineno}
\usepackage{comment}
\usepackage{hyperref}
\usepackage{url}
\usepackage{graphicx}
\usepackage{textgreek}
\usepackage{multirow}
\usepackage{caption}
\usepackage{subcaption}
\usepackage{tablefootnote}
\usepackage[table]{xcolor}
\usepackage{bm}
\usepackage[free-standing-units=true]{siunitx}
\usepackage[normalem]{ulem}
\usepackage{booktabs}

\newcommand{\uproman}[1]{\uppercase\expandafter{\romannumeral#1}}

\title{\boldmath AI-based particle track identification in scintillating fibres read out with imaging sensors}

\author[1]{Noemi Bührer}
\author[1]{Saúl Alonso-Monsalve}
\author[1]{Matthew Franks}
\author[1]{Till Dieminger}
\author[1]{Davide Sgalaberna}
\affiliation[1]{ETH Zürich,\\
Institute for Particle Physics and Astrophysics, CH-8093 Zurich, Switzerland.}

\emailAdd{nritzmann@ethz.ch}

\abstract{
This paper presents the development and application of an AI-based method for particle track identification using scintillating fibres read out with imaging sensors. We propose a variational autoencoder (VAE) to efficiently filter and identify frames containing signal from the substantial data generated by SPAD array sensors. Our VAE model, trained on purely background frames, demonstrated a high capability to distinguish frames containing particle tracks from background noise. The performance of the VAE-based anomaly detection was validated with experimental data, demonstrating the method's ability to efficiently identify relevant events with rapid processing time, suggesting a solid prospect for deployment as a fast inference tool on hardware for real-time anomaly detection. This work highlights the potential of combining advanced sensor technology with machine learning techniques to enhance particle detection and tracking.
}


\begin{document}
\maketitle
\flushbottom

\section{Introduction}
\label{sec:intro}

Scintillating fibre (SciFi) detectors are among the innovative concepts currently being explored for future particle detection. Offering excellent spatial resolution, these detectors are critical for enhancing the reconstruction of low-momentum particles~\cite{Collaboration:1647400}.
However, constructing large detector volumes with SciFi would lead to an incredibly high number of channels if read out with silicon photomultipliers (SiPMs).
Hence, different SciFi readout schemes based on imaging sensors have been used. For example, in Ref.~\cite{K2K:2000kji}, CCD cameras were coupled with image intensifier tubes.
Recent work demonstrated particle tracking with SciFi read out by a CMOS single-photon avalanche diode (SPAD) array sensor \cite{franks2023demonstration}, illustrating the feasibility of reading out a SciFi volume with single fibre resolution and minimising the number of electronics channels.
However, the significant data volume and the challenge of distinguishing frames containing signal from frames containing background (i.e., dark counts or crosstalk) remain issues.
As a form of anomaly detection, machine learning can be used to learn from background samples alone, identifying signal events based on their deviation from the background. This approach is more robust than traditional methods that require a predefined mix of background and signal samples for training, and it is typically much faster compared to traditional solutions~\cite{Belis_2024}. In this paper, we present an AI-based method using a variational autoencoder (VAE) to identify particle tracks in scintillating fibres read out with a SPAD array sensor. The method has been tested on the data acquired in Ref.~\cite{franks2023demonstration}. The VAE is trained exclusively on background frames to learn the typical noise pattern, enabling it to identify frames containing signal events with higher accuracy. We evaluate the performance of our model using experimental data and demonstrate its ability to distinguish between signal and background frames effectively. 

The rest of the article is structured as follows: Section~\ref{sec:autoencoders} provides a brief overview of autoencoders and their application in anomaly detection. Section~\ref{sec:anomaly_detection_vae} presents the VAE architecture, training process, and methodology for anomaly detection; it also discusses the results of applying the VAE to experimental data, including evaluating its computational performance. Finally, Section~\ref{sec:conclusion} concludes the paper and outlines potential future work.

\section{Autoencoders}
\label{sec:autoencoders}

An autoencoder is a self-supervised neural network that is trained to reconstruct an output which is close to its original input. It consists of two parts, an encoder and a decoder. The encoder $f$ transforms the input $\bm{x}$ into a latent representation $\bm{z} = f(\bm{x})$ and the decoder maps the latent representation $\bm{z}$ onto the reconstructed output $\bm{\hat{x}} = g(\bm{z})$. The difference between the input $\bm{x}$ and the reconstruction $\bm{\hat{x}}$ is called reconstruction error. The autoencoder learns to minimise this error. Thereby, the main goal is not for the autoencoder to reconstruct the input perfectly, but that it learns a useful representation of the data. This can be achieved by introducing some form of constraint, for example by forcing $\bm{z}$ to have lower dimension than $\bm{x}$~\cite{an2015variational,Bishop2024,Goodfellow-et-al-2016}.

The concept of autoencoders has been generalised from deterministic encoder and decoder functions to stochastic models which learn encoder and decoder distributions. These models are called variational autoencoders~\cite{Bishop2024}.

\subsection{Variational Autoencoders}

Just like the standard autoencoder, a VAE is composed of an encoder and a decoder which are trained to minimise a reconstruction error. However, as opposed to the standard autoencoder where the input is encoded as a single point, the VAE encodes the input as a distribution over the latent space, allowing the model to capture uncertainty in the data and creating a continuous and smooth latent space, among other benefits.

In more detail, the latent variable $\bm{z}$ of some input $\bm{x}$ is constrained to be a random variable which is distributed according to some prior distribution $p_{\bm{\theta}}(\bm{z})$. The true posterior distribution $p_{\theta}(\bm{z}|\bm{x})$ is intractable but it can be approximated by $q_{\bm{\phi}}(\bm{z}|\bm{x})$ using variational inference techniques. This sets up a family of parameterised distributions as priors, usually Gaussian distributions with mean 0 and unitary variance, and chooses the distribution with the smallest approximation error. The parameters are optimised according to
\begin{equation}
    \mathcal{L}(\bm{\theta}, \bm{\phi}, \bm{x}) = \mathbb{E}_{q_\Phi(\bm{z}|\bm{x})}[p_\theta(\bm{x}|\bm{z})] - \mathcal{D}_{KL}(q_\Phi(\bm{z}|\bm{x}) || p_\theta(\bm{z})).
\end{equation}
The first term is equivalent to the reconstruction error and measures the reconstruction quality. The second term is the Kullback-Leibler Divergence (KLD). This ensures that the approximation $q_{\bm{\phi}}(\bm{z}|\bm{x})$ is close to the true posterior distribution $p_{\theta}(\bm{z}|\bm{x})$. \cite{schneider, bergamin2022novel}. 

\begin{figure}[ht]
    \centering
    \includegraphics[width=0.5\linewidth]{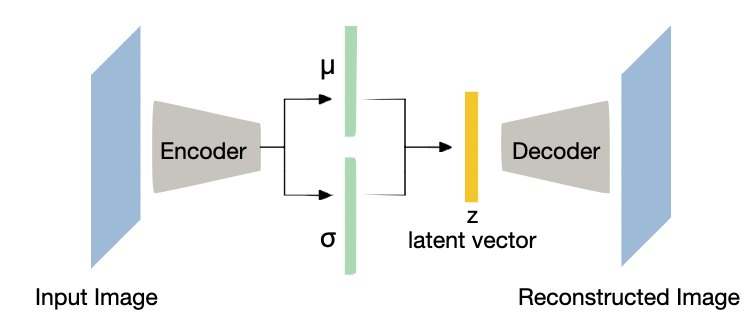}
    \caption{Variational Autoencoder simplified structure.}
    \label{fig:sketch_vae}
\end{figure}

In practice, this means, given an input $\bm{x}$, the network returns the parameters defining the probability distribution $q_{\bm{\phi}}(\bm{z}|\bm{x})$ over the latent space that is most likely to produce $\bm{x}$. Since this distribution was constrained to be Gaussian, the defining parameters are the mean $\bm{\mu_{\phi}}(\bm{x})$ and the variance $\bm{\sigma_{\phi}^2}(\bm{x})$~\cite{bergamin2022novel}. The latent variable $\bm{z}$ is then sampled from $q_{\bm{\phi}}(\bm{z}|\bm{x})$. However, backpropagation cannot deal with this random sampling. Therefore, the sampling is done via a reparameterisation trick by describing $\bm{z}$ as a differentiable function of another random variable $\varepsilon \sim \mathcal{N}(0, \mathrm{I})$. Then the latent variable is calculated as $\bm{z} = \bm{\mu_{\phi}}(\bm{x}) + \bm{\sigma_{\phi}}(\bm{x})\odot \bm{\varepsilon}$ where $\odot$ is the Hadamard product~\cite{schneider}. The decoder reconstructs the input using the sampled $\bm{z}$ as for a standard autoencoder. This workflow is illustrated in Fig.~\ref{fig:sketch_vae}.

\subsection{Anomaly Detection with VAE}
VAEs have shown success in various anomaly detection tasks \cite{10.1145/3097983.3098052,bergamin2022novel, 10.1145/3178876.3185996, lindenbaum2022probabilisticrobustautoencodersoutlier}. As discussed at the beginning of this section, autoencoders map an input into a lower dimensional latent space. Through this constraint, it is expected, that the autoencoder learns useful features of the data. It learns to reconstruct data frequently exposed during training but fails on rare instances in the training set. Thus, when training an autoencoder on a dataset, the reconstruction is expected to fail when encountering anomalous data~\cite{an2015variational}. 

However, the connection between anomalies and large reconstruction errors is not entirely clear~\cite{8999265}. Using VAEs has the advantage that the KLD term regularises the latent space by forcing it to follow a well-defined prior distribution. The model learns the distribution of latent variables instead of the variables itself. This means, we can label an event as anomalous if the probability of it being associated with the learned latent distribution is low. Therefore, instead of the reconstruction error, also the regularised latent space can be used for anomaly detection \cite{Belis_2024}.

\subsection{Anomaly Detection in Particle Physics}

In collider experiments, anomaly detection is typically used to monitor detectors and identify unexpected events, potentially indicating new physics beyond the Standard Model (BSM). Machine learning methods, including VAEs, enable model-agnostic anomaly detection, enhancing the discovery potential for BSM events~\cite{Cerri2019, fraser2022challenges, Belis_2024}.

Recent advancements have shown that VAEs can be particularly effective in high-dimensional data typical of particle physics experiments~\cite{liu2023fastparticlebasedanomalydetection, 10.1007/978-3-031-51023-6_14}. By learning the underlying distribution of the Standard Model events, VAEs are hypothesised to be capable of distinguishing between common physical processes and rare anomalies that may signal new physics. For instance, studies have demonstrated the ability of VAEs to identify anomalies in the Large Hadron Collider (LHC) data, aiding in the search for novel particles or interactions~\cite{10.21468/SciPostPhys.12.1.045, PhysRevD.107.016002}. 

Future research may focus on enhancing the interpretability of the latent space in VAEs, facilitating a deeper understanding of the detected anomalies. Additionally, integrating VAEs with other machine learning approaches, such as supervised learning algorithms, could further improve anomaly detection capabilities. The development of more sophisticated models and training techniques will likely continue to expand the applicability of VAEs in particle physics and other scientific domains~\cite{Belis_2024}.

\section{Anomaly Detection with VAE}
\label{sec:anomaly_detection_vae}

This section details the datasets along with a standard method for identifying tracks. Besides, the VAE model architecture, training process, and its application for anomaly detection are also presented.

\subsection{Experimental Setup and Data} 

\begin{figure*}
    \centering
    \begin{subfigure}{0.44\textwidth}
        \centering
        \includegraphics[width=\linewidth]{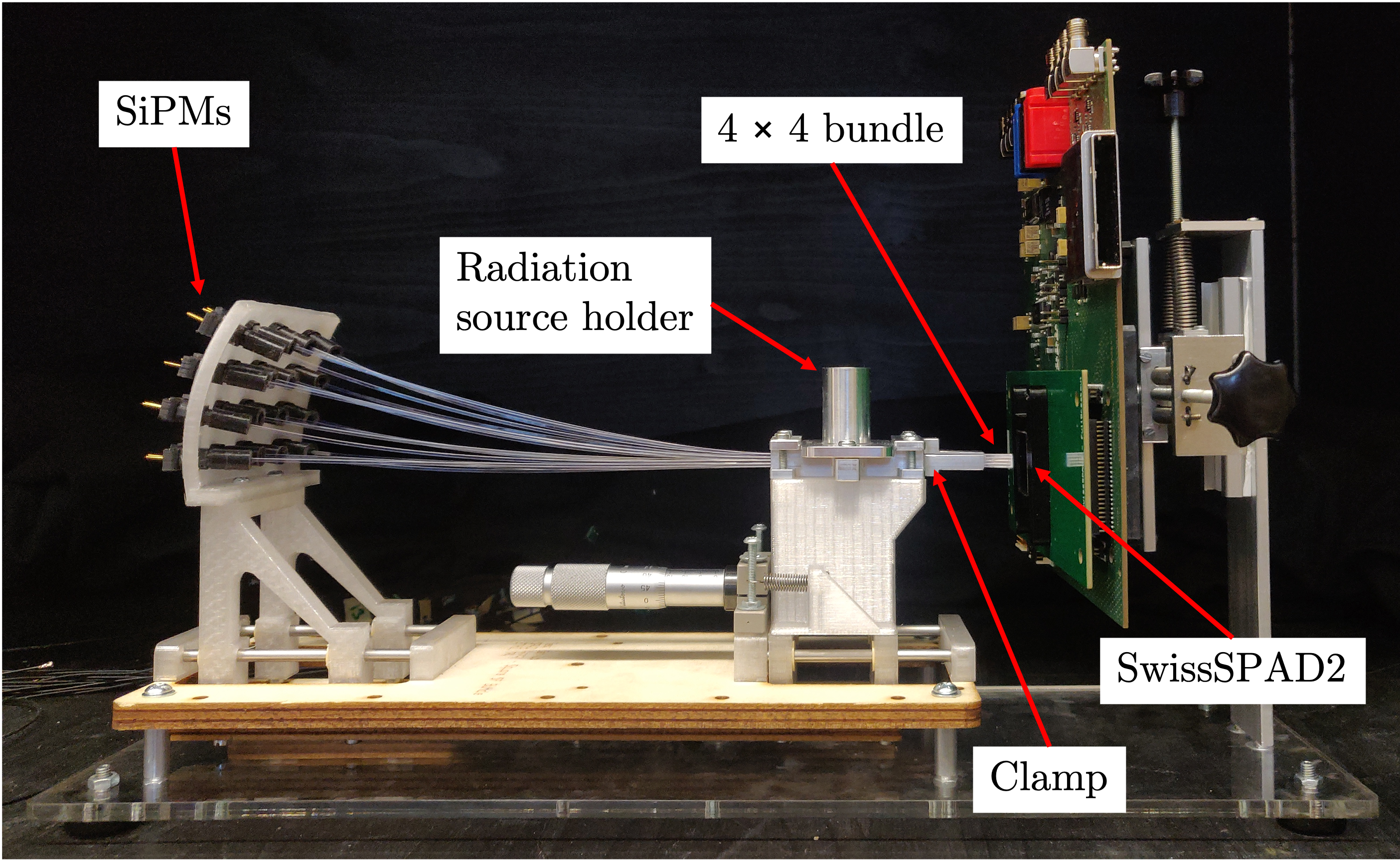}
        \caption{}
        \label{fig:SciFi_setup}
    \end{subfigure}%
    \hspace{0.001\textwidth}
    \begin{subfigure}{0.27\textwidth}
        \centering
        \includegraphics[width=\linewidth, clip, trim=5cm 5cm 5cm 5cm]{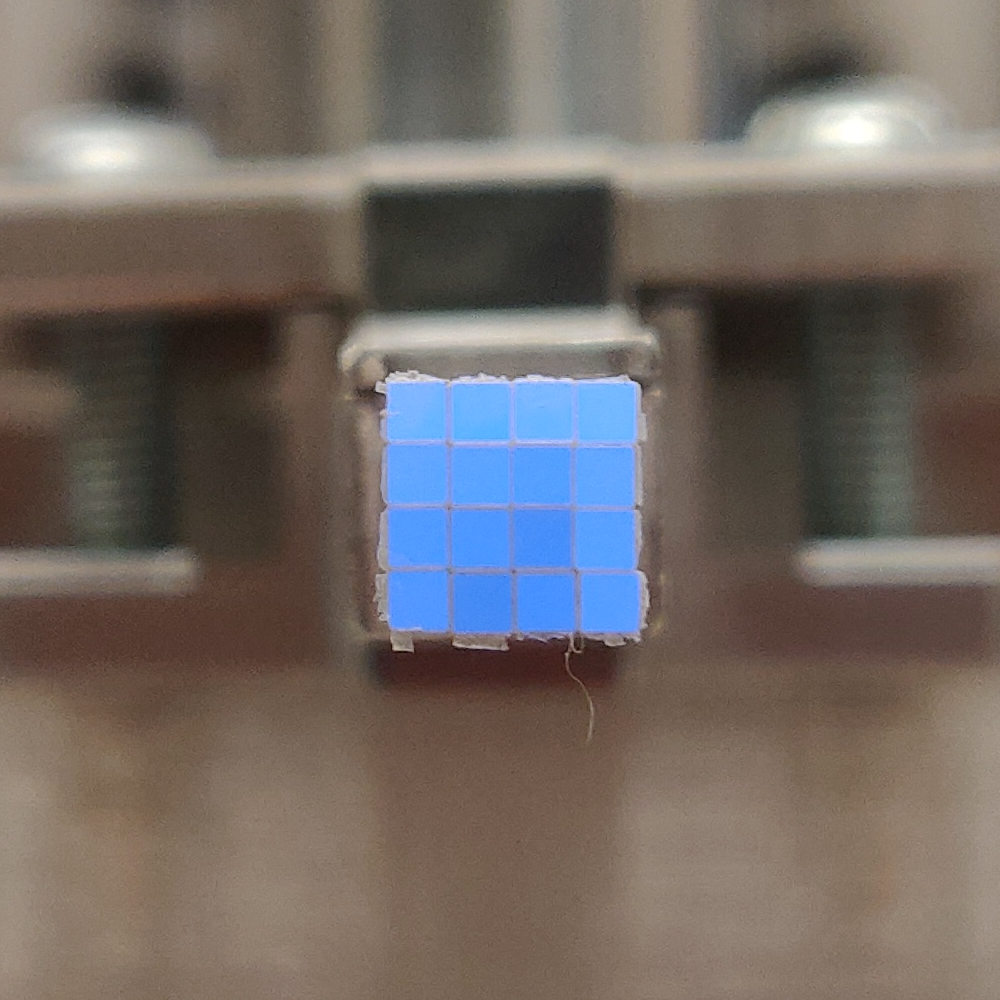}
        \caption{}
        \label{fig:SciFi_bundle}
    \end{subfigure}%
    \hspace{0.001\textwidth}
    \begin{subfigure}{0.27\textwidth}
        \centering
        \includegraphics[width=\linewidth, clip, trim=10cm 10cm 10cm 10cm]{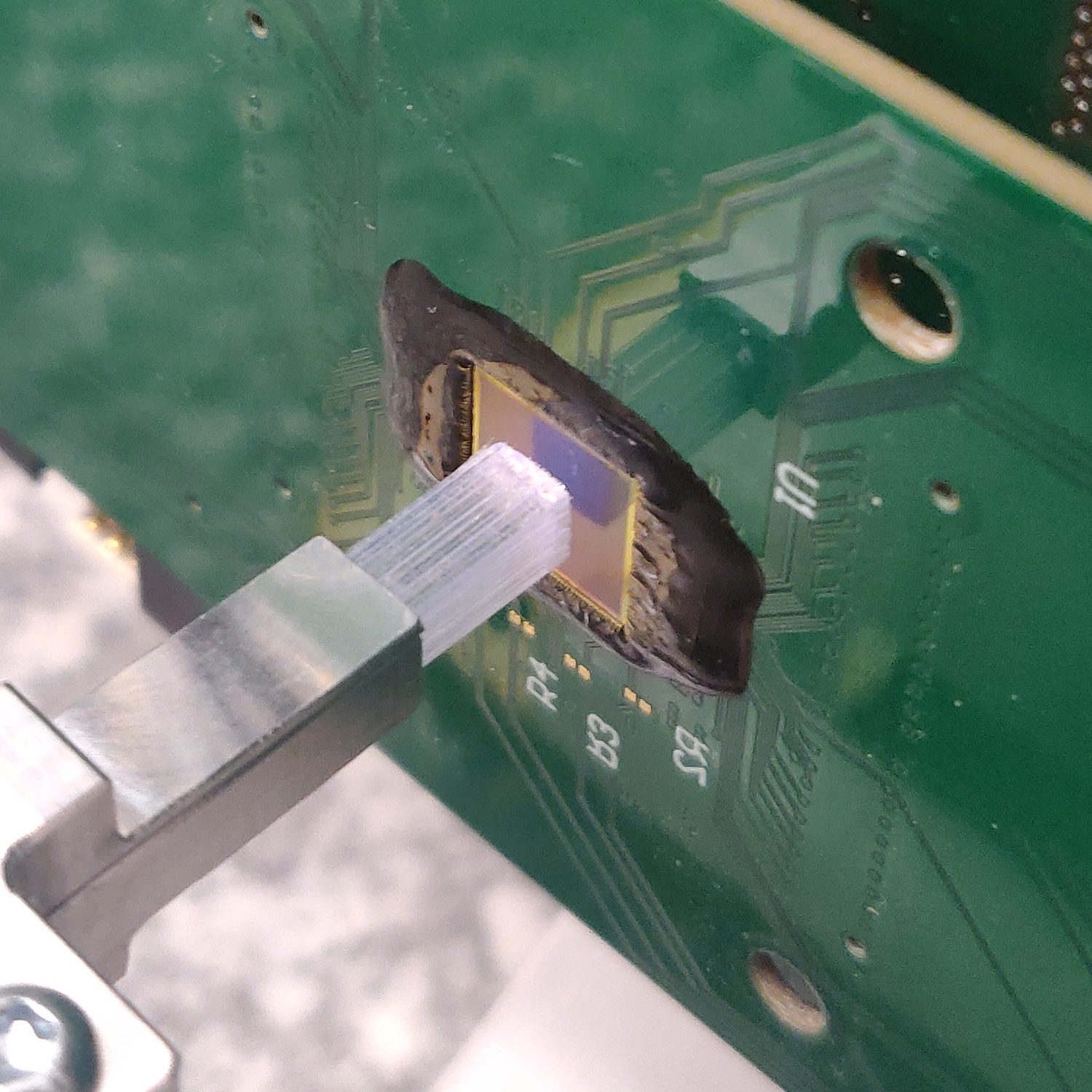}
        \caption{}
        \label{fig:SciFi_setup_zoom}
    \end{subfigure}
    \caption{
    (a) SciFi setup used for the acquisition of the data samples. Scintillating fibres are independently coupled to SiPM on their left side end, while on the opposite side they are bundled together and coupled to a single SwissSPAD2 sensor.
    (b) Close-up photograph of the face of $4\times4$ scintillating fibre bundle.
    (c) Coupling of the scintillating fibre bundle with SwissSPAD2. Black glob top was used to protect the wire bonding around the active area. Taken from~\cite{franks2023demonstration}.
    }
    \label{fig:scifi-prototype}
\end{figure*}

The dataset analysed in this study is the same as in Ref.~\cite{franks2023demonstration}. 
Sixteen square scintillating fibres~\cite{datasheet_plastic_scint_kurraray} with a diameter of 1~mm were arranged into a 4\texttimes4 bundle and coupled to SwissSPAD2~\cite{Ulku_2019_A_512_x}.
On top of the bundle of fibres, a 185~kBq-$^{90}$Sr source could be placed; see Fig.~\ref{fig:scifi-prototype}.
The electrons from the source then travel through the fibres, creating scintillation photons, which are subsequently detected by the SPAD. 
The setup was placed in a dark chamber to limit noise due to stray light. 

A first sample, without the $^{90}$Sr source present, subsequently called the background ($\mathrm{BG}$) sample, consists of \num[group-separator={,}]{5012736} frames. 
A single frame contains all detected counts in a 1~$\mu s$ time window.
Counts in these frames are due to sensor noise, known as dark counts. 
A second sample, this time with the $^{90}$Sr source placed centrally on top of the fibres, contains the same number of frames and is subsequent called the signal sample ($^{90}\mathrm{Sr+BG}$). 
Counts in this sample can contain both dark counts and detected scintillation photons. 

\begin{figure}
    \centering
    \includegraphics[width=0.78\textwidth]{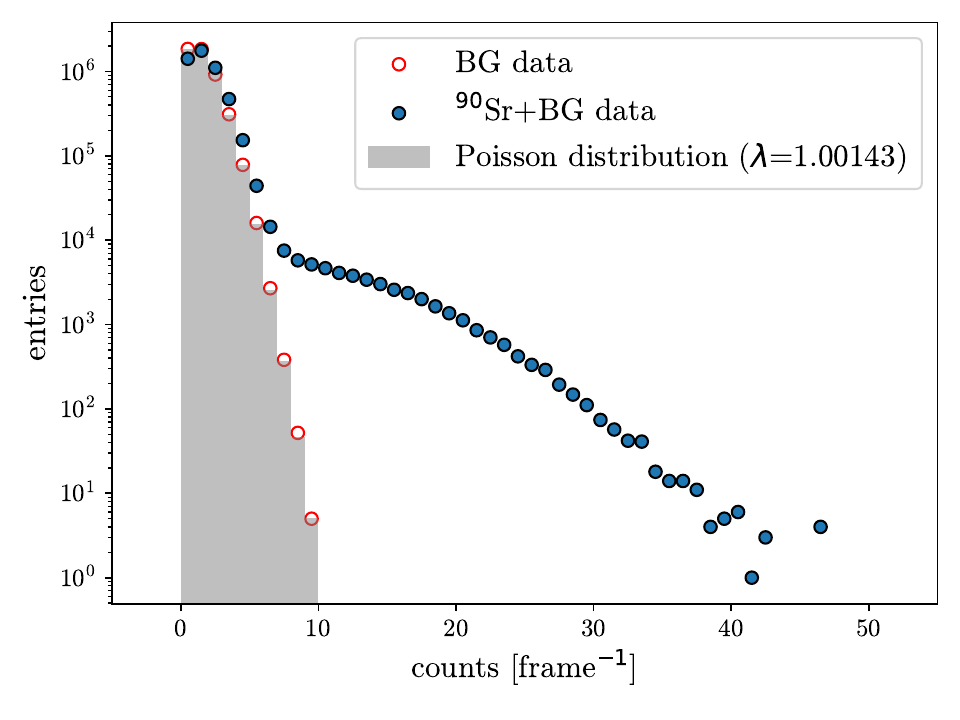}
    \caption{Distribution of number of counts per 1~$\mu s$ frame corresponding to the fibre bundle region on SwissSPAD2. 
    The hollow red markers show the distribution obtained from data collected without $^{90}$Sr using SwissSPAD2, known as the background ($\mathrm{BG}$) data. The grey histogram is a Poisson distribution with $\lambda$ equal to mean counts of the $\mathrm{BG}$ data. The solid blue dots show the distribution of the number of counts per frame taken with the $^{90}$Sr source positioned above the fibre bundle. Adjusted from~\cite{franks2023demonstration}.}
    \label{fig:count_distribution}
\end{figure}

In~\cite{franks2023demonstration}, a simple cutoff in the number of detected counts was used to differentiate between signal- and background-like frames in the $^{90}\mathrm{Sr+BG}$ sample. 
The cutoff used was ten counts, the highest number of counts detected in the background sample. 
Figure~\ref{fig:count_distribution} shows that in the intermediate region between five and ten counts, potential signal frames are present. 
Identifying signal-like frames in this region requires more complex analysis steps.

\subsection{Standard Method for Identifying Electron Tracks}
\label{sec:standard_method}

\begin{figure}
    \centering
    \includegraphics[width=0.9\textwidth]{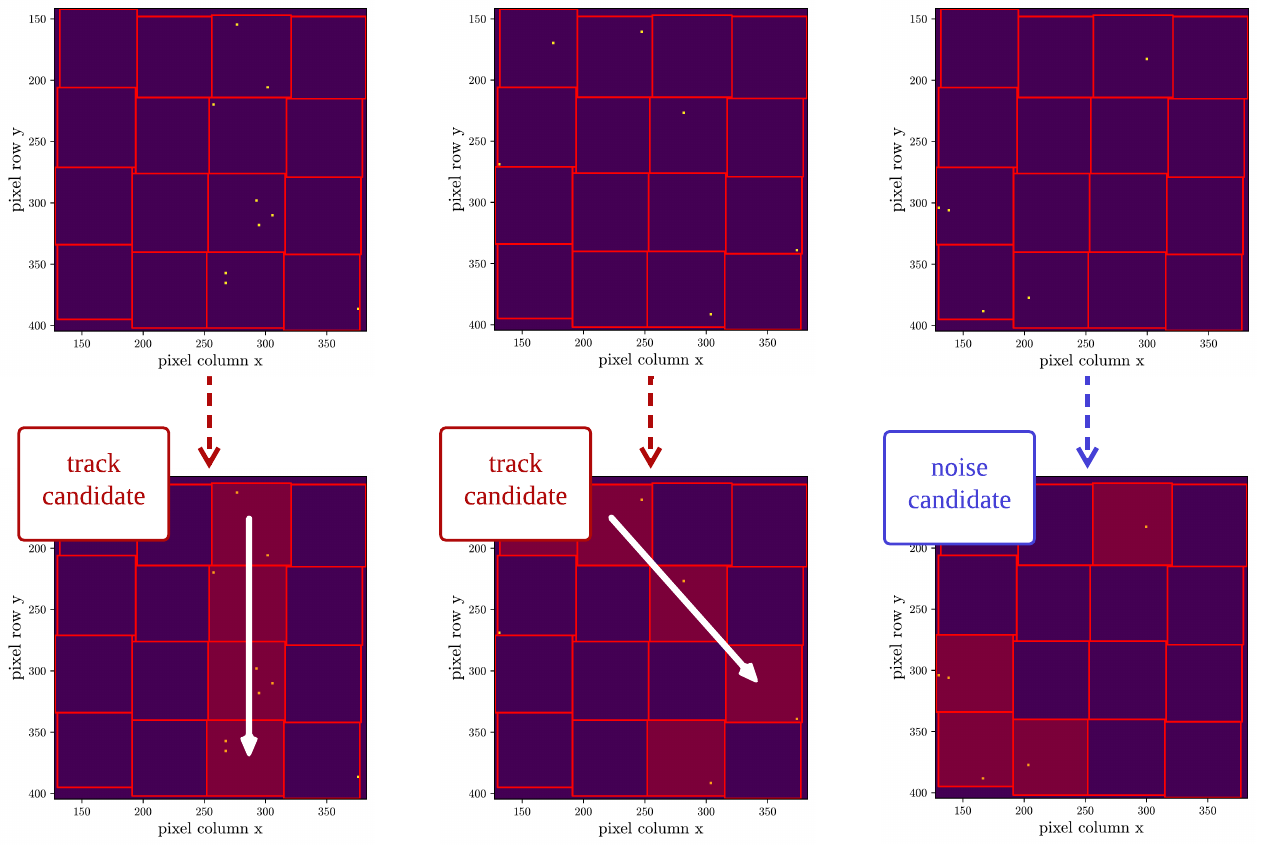}
    \caption{The standard track selection process, following the approach described in Ref.~\cite{franks2023demonstration}, is illustrated for three example frames. The red lines indicate the fibre contours, while the filled squares highlight fibres with at least one photon count. The left frame is identified as a signal event, as it contains detected photons on at least three vertically down-going fibres. Similarly, the central frame is classified as signal due to the presence of a diagonal down-going track. In contrast, the right frame is categorised as noise, as it lacks a discernible down-going pattern.
}
    \label{fig:standard_method}
\end{figure}

Due to the lack of available studies performing systematic track selection for scintillating fibre detectors read out by imaging sensors, we describe a standard method for identifying electron tracks inspired by the approach used in Ref.~\cite{franks2023demonstration}. This reference method is designed to detect vertical track-like events using pixel-to-fibre mapping, requiring electrons to traverse a minimum number of fibres while accumulating a specified number of counts per track and per fibre (see Fig.~\ref{fig:standard_method}). This selection strategy leverages the expected behaviour of electrons from a $^{90}$Sr source as they interact with the scintillating fibres, producing detectable photon signals. To define a valid track, the method enforces a minimum count threshold per fibre and requires a continuous trajectory across multiple fibres. Frames that fail to meet these criteria are discarded as background events. This method aims to distinguish true electron-induced signals from random noise and dark counts, thereby improving the reliability of track identification in experimental data.

It is important to note that this selection method was originally applied to the entire dataset, which includes frames with a wide range of detected counts, often exceeding 10. The effectiveness of the approach was primarily demonstrated in high-count regions, where track identification is more reliable due to the increased signal-to-noise ratio.

\subsection{VAE Architecture and Training Details}

The VAE used in this study comprises an encoder with six convolutional layers, each followed by LeakyReLU activations with a negative slope of 0.01. The convolutional layers progressively increase from 8 to 64 filters, using $4 \times 4$ kernels with a stride of 2, effectively downsampling the input while extracting features. The encoder output is flattened and passed through fully connected layers to produce the mean ($\bm{\mu}$) and log-variance ($\log(\bm{\sigma}^2)$) vectors for the 32-dimensional latent space. The decoder uses a symmetrical structure with six transposed convolutional layers and LeakyReLU activations to upsample the latent vector back to the original input dimensions, with a final sigmoid activation ensuring output values in the range [0, 1]. The model contains 605,513 trainable parameters, balancing complexity with computational efficiency.

The loss function $\mathcal{L}$ combines the Binary Cross-Entropy (BCE) loss and KLD.

The BCE loss is given by:
\begin{equation}
    \mathcal{L}_{\text{BCE}}(\bm{x}, \bm{\hat{x}}) = -\sum_{n}\left[x_n \cdot \log(\hat{x}_n) + (1 - x_n) \cdot \log(1 - \hat{x}_n)\right]
\end{equation}
where $x_n$ represents the ground truth binary label for the $n$-th pixel, and $\hat{x}_n$ is the predicted probability that the label is $1$. Here, $\bm{x}$ and $\bm{\hat{x}}$ are vectors containing the ground truth labels and predicted probabilities for all pixels, respectively.

The KLD term is:
\begin{equation}
    \mathcal{L}_{\text{KLD}}(\bm{\mu}, \bm{\sigma}) = -\frac{1}{2}\sum_{i}\left[1 + \log(\sigma_i^2) - \sigma_i^2 - \mu_i^2 \right]
\end{equation}
where $\bm{\mu} = (\mu_1, \mu_2, \dots, \mu_d)$ and $\bm{\sigma} = (\sigma_1, \sigma_2, \dots, \sigma_d)$ are the mean and standard deviation vectors of the latent variable's Gaussian distribution. The summation is over the latent space dimensions, with \( d \) being the dimensionality of the latent space.

The overall loss function $\mathcal{L}$ is then a combination of both terms:
\begin{equation}
    \mathcal{L}(\bm{x}, \bm{\hat{x}}, \bm{\mu}, \bm{\sigma}) = \mathcal{L}_{\text{BCE}}(\bm{x}, \bm{\hat{x}}) + \beta \cdot \mathcal{L}_{\text{KLD}}(\bm{\mu}, \bm{\sigma})
\end{equation}
where $\beta$ is a weighting factor applied to the KLD term. In this work, the $\beta$ term is introduced following the approach outlined by~\cite{fu2019cyclicalannealingschedulesimple}, where it is scheduled using a sigmoid function during a 5-cycle annealing schedule~\cite{bowman2016generatingsentencescontinuousspace}. This scheduling mitigates the issue of KLD vanishing and facilitates the formation of a well-organised latent space.

\begin{figure}[ht]
    \centering
    \includegraphics[width=0.7\textwidth]{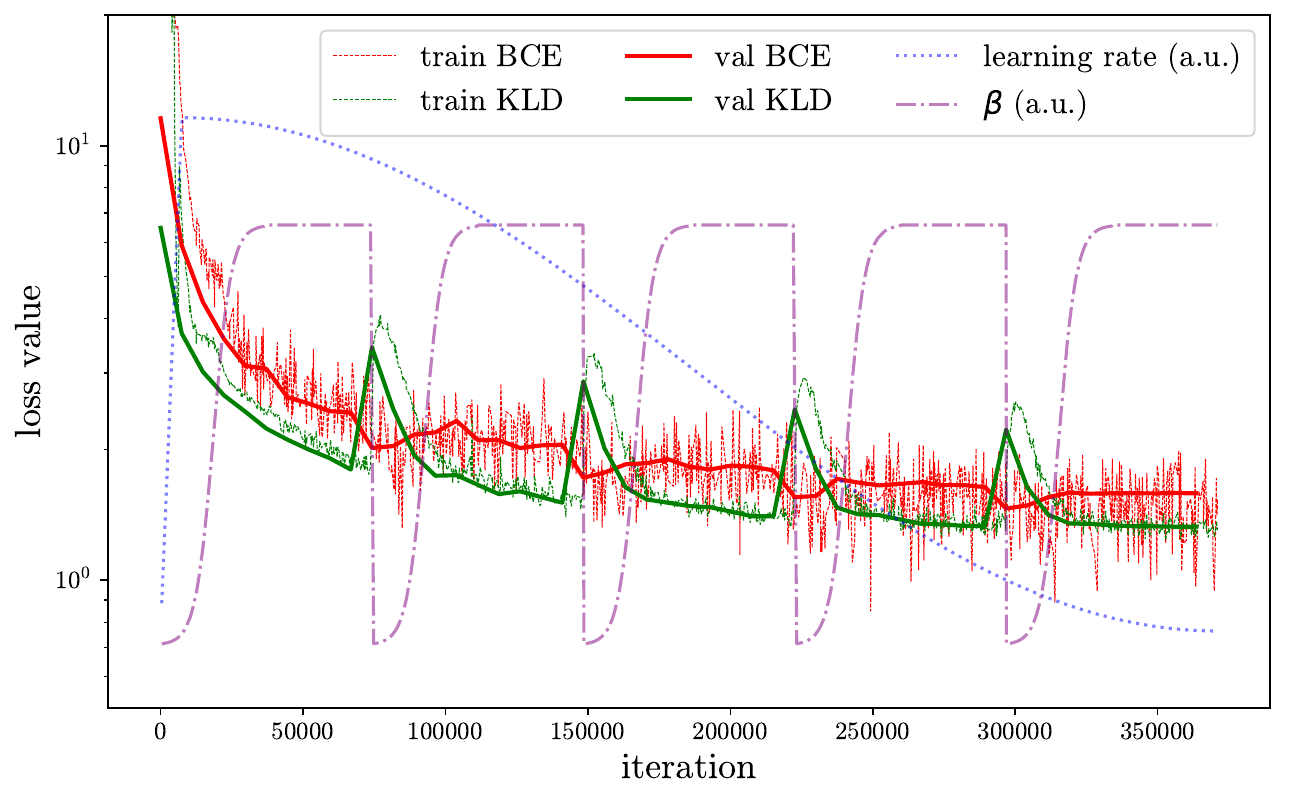}
    \caption{The plot illustrates the training and validation losses of a Variational Autoencoder (VAE) across iterations, with Binary Cross-Entropy (BCE) loss shown in red and Kullback-Leibler Divergence (KLD) loss in green. Solid lines represent validation losses, while dashed lines denote training losses. The learning rate (blue dotted line) and the $\beta$ weight of the KLD term (purple dash-dot line) are also plotted in arbitrary units.}
    \label{fig:loss_func_6e-1}
\end{figure}

The VAE was implemented using PyTorch 1.14.0~\cite{Ansel_PyTorch_2_Faster_2024} with Python 3.8.10 and trained on an NVIDIA V100 GPU with a batch size of 256 for 50 epochs. The model parameters were optimised using the AdamW optimiser with a learning rate of $1 \times 10^{-3}$. To ensure a smooth training process, we employed a learning rate schedule that consisted of a linear warm-up phase followed by cosine annealing. The warm-up phase was applied during the first epoch, after which a cosine annealing schedule was used for the remaining epochs. This combined scheduling approach facilitated stable initial convergence while enabling efficient exploration of the loss landscape during later stages of training. The $\mathrm{BG}$ dataset was split in the following way: $60\%$ for training, $10\%$ for validation, and $30\%$ for testing. The training and validation losses are shown in Fig.~\ref{fig:loss_func_6e-1}.

\subsection{Anomaly Detection}
\label{sec:anomaly_detection}

\begin{figure}[htbp]
    \centering
    \begin{subfigure}{0.49\textwidth}
        \centering
        \includegraphics[width=1.\linewidth]{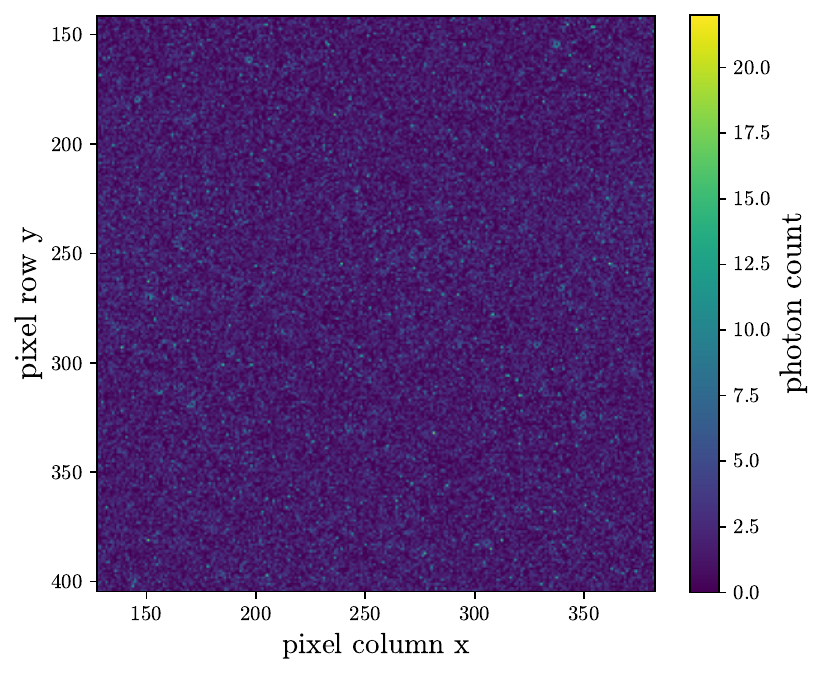}
        \caption{$\mathrm{BG}$ (test set).}
        \label{fig:sum_frames_all_bkg}
    \end{subfigure}
    \hfill
    \begin{subfigure}[b]{0.49\textwidth}
        \centering
        \includegraphics[width=1.\linewidth]{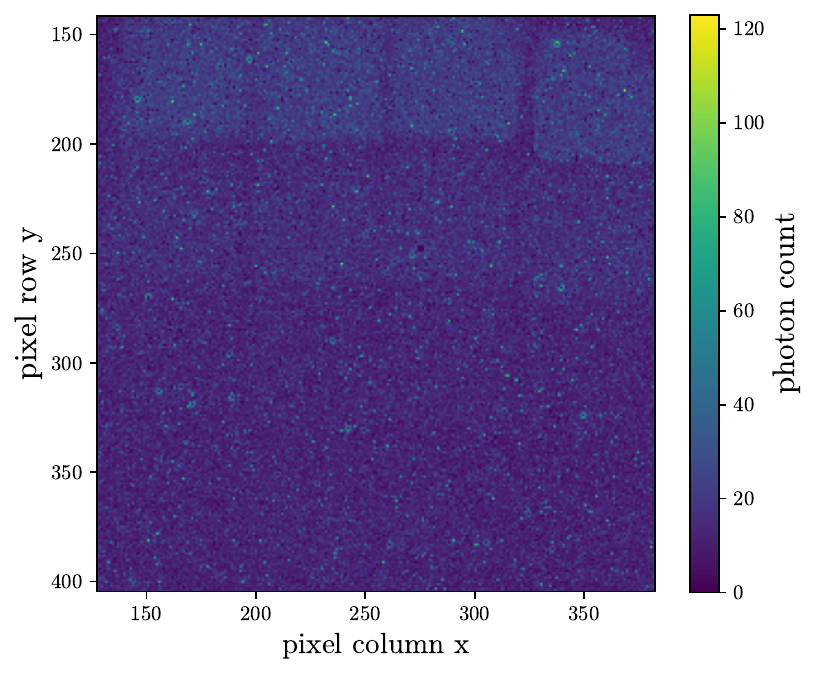}
        \caption{$^{90}\mathrm{Sr+BG}$.}
        \label{fig:sum_frames_all_sig}
    \end{subfigure}
    \caption{Total frame sum within the fibre bundle region. Data taken from Ref.~\cite{franks2023demonstration}}
    \label{fig:sum_frames_all}
\end{figure}

Although the VAE is trained on the entire BG dataset, the anomaly detection analysis specifically targets frames with four to ten counts. This range is chosen to maximise effectiveness because it represents the photon count where distinguishing the presence of a signal remains uncertain. Visually detecting signal tracks is extremely challenging due to the low photon counts per frame, making it difficult to distinguish them from background events that only contain DCs. However, by examining the total sum of frames for both the $\mathrm{BG}$ test set and the $^{90}\mathrm{Sr+BG}$ dataset (Fig.~\ref{fig:sum_frames_all}), we can slightly identify electrons traversing the scintillating material (from top to bottom in Fig.~\ref{fig:sum_frames_all_sig}). In contrast, the image in the BG case appears as pure noise (Fig.~\ref{fig:sum_frames_all_bkg}).

\begin{figure}[htbp]
    \centering
    \begin{subfigure}{0.49\textwidth}
        \centering
        \includegraphics[width=1.\linewidth]{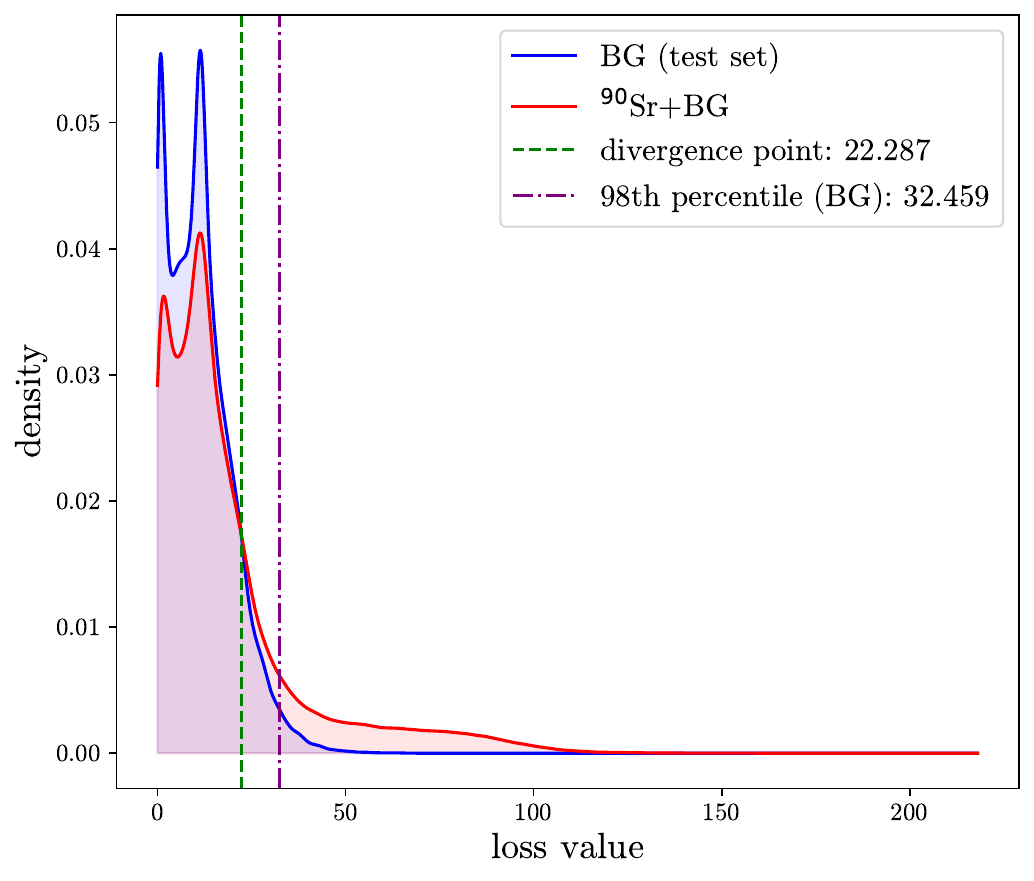}
        \caption{BCE loss.}
        \label{fig:losses_comparison_bce}
    \end{subfigure}
    \hfill
    \begin{subfigure}[b]{0.49\textwidth}
        \centering
        \includegraphics[width=1.\linewidth]{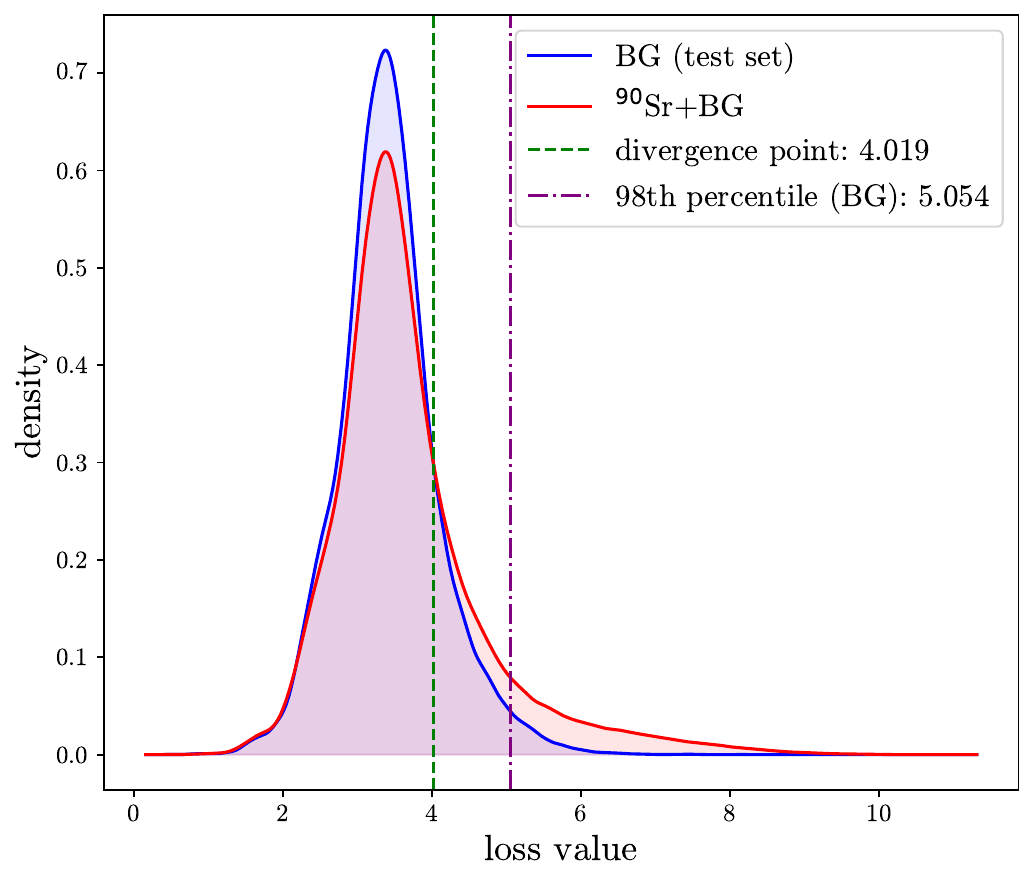}
        \caption{KLD loss.}
        \label{fig:losses_comparison_kld}
    \end{subfigure}
    \caption{Comparison of loss distributions for the Variational Autoencoder (VAE) model. The plot on the left shows the distribution of the Binary Cross-Entropy (BCE) reconstruction loss, while the right shows the distribution of the Kullback-Leibler Divergence (KLD) loss. In both plots, the blue curve represents the loss distribution for the $\mathrm{BG}$ test set, and the red curve represents the loss distribution for the $^{90}\mathrm{Sr+BG}$ data. The vertical dashed green lines indicate the divergence points, where the loss distributions of the $^{90}\mathrm{Sr+BG}$ data begin to exceed those of the $\mathrm{BG}$ test set. The vertical dashed-dotted purple lines mark the 98th percentile of the loss distribution for the $\mathrm{BG}$ test set, serving as a threshold to identify outliers or regions of interest where the loss values are notably high. The distributions are computed by evaluating the BCE and KLD losses for each individual frame in the dataset. The x-axis represents the per-frame loss values, and the y-axis represents the probability density function (PDF) estimated using Kernel Density Estimation (KDE). This visualisation allows us to identify points where the loss values of the $^{90}\mathrm{Sr+BG}$ dataset start deviating from the $\mathrm{BG}$ test set, motivating the selection of thresholds for anomaly detection.}
    \label{fig:losses_comparison}
\end{figure}

To identify such signal frames in the $^{90}\mathrm{Sr+BG}$ dataset, we utilise the loss values from our VAE. Figure~\ref{fig:losses_comparison} illustrates the distribution of BCE and KLD loss values for the test set of the $\mathrm{BG}$ and the $^{90}\mathrm{Sr+BG}$ dataset. The plots indicate the divergence points, where the loss distributions of the $^{90}\mathrm{Sr+BG}$ data start to exceed those of the BG test set. Both these divergence points and the 98th percentile of the $\mathrm{BG}$ test set, which is also marked in the plots, are used to identify anomalies. The 98th percentile was chosen empirically to ensure anomalies detected with this threshold are well beyond the divergence point while balancing signal retention and background rejection. Tests with alternative thresholds from the 95th percentile yielded consistent results, demonstrating the method's robustness to small variations in the cutoff.

\begin{figure}[htbp]
    \centering
    \begin{subfigure}{0.99\textwidth}
        \centering
        \includegraphics[width=1.\linewidth]{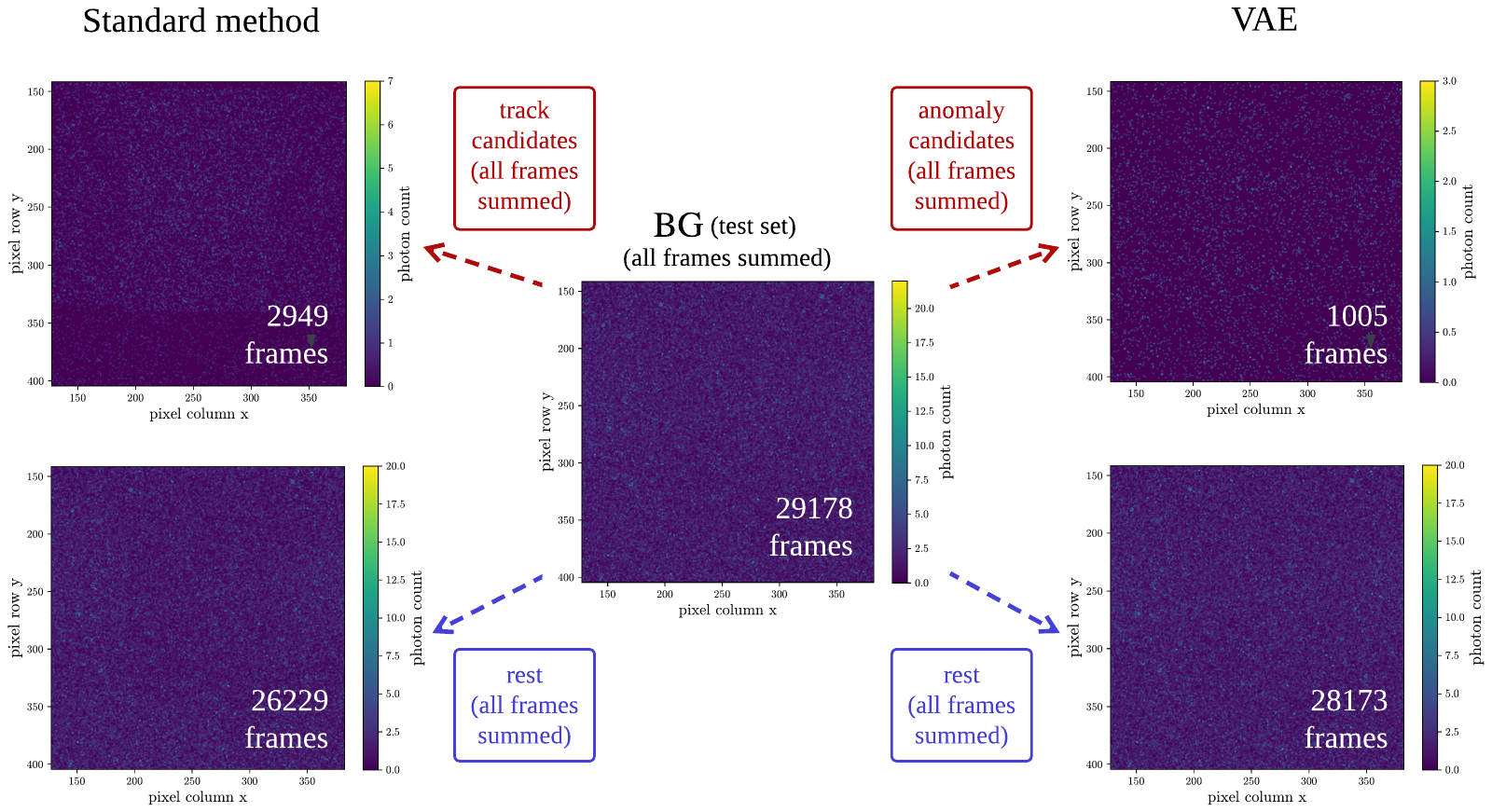}
        \caption{Anomaly selection on the test set of the $\mathrm{BG}$ sample.}
        \label{fig:sum_frames_bkg}
    \end{subfigure}
    \begin{subfigure}[b]{1.0\textwidth}
        \centering
        \includegraphics[width=1.\linewidth]{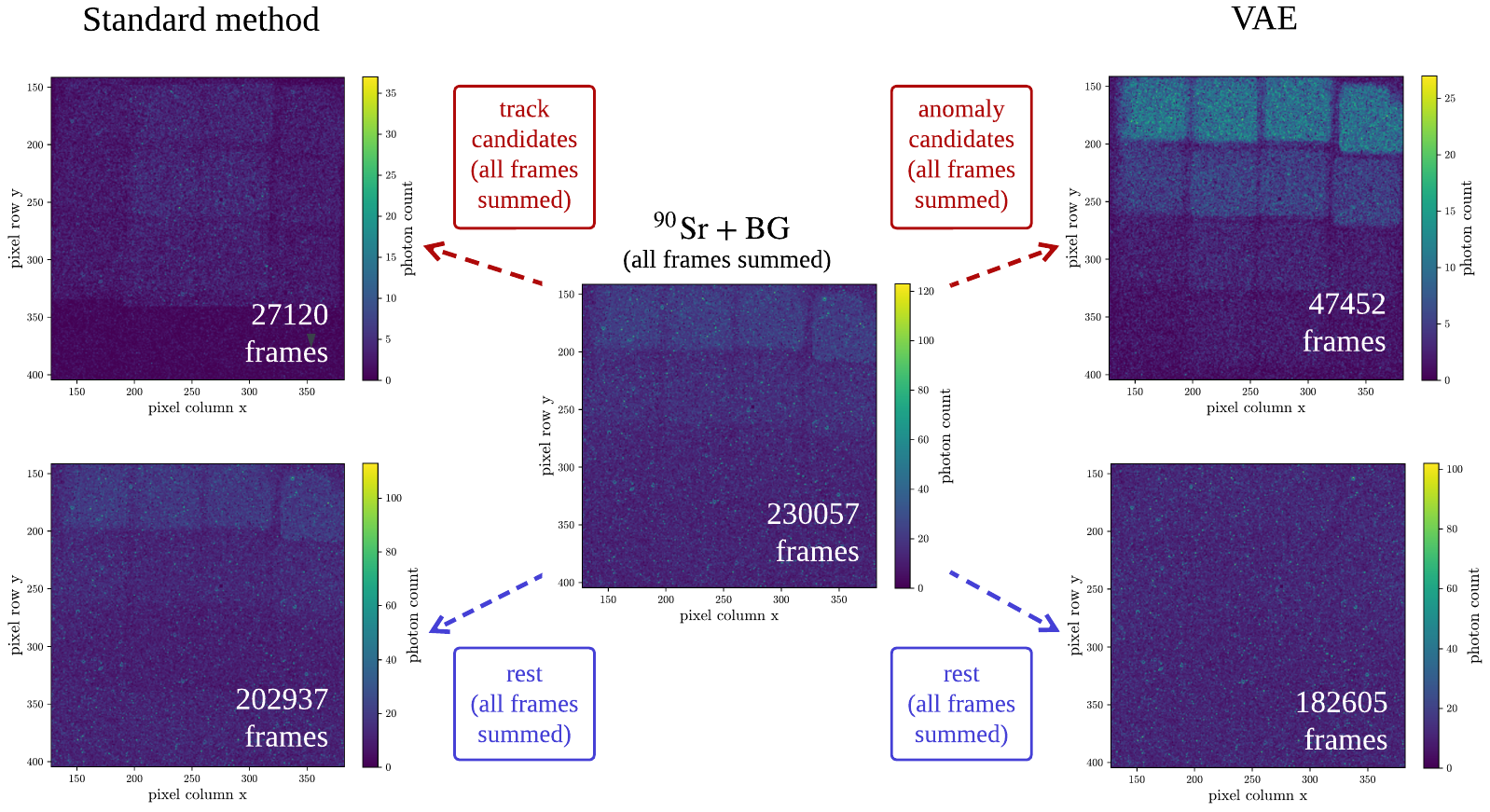}
        \caption{Anomaly selection on the $^{90}\mathrm{Sr+BG}$ sample.}
        \label{fig:sum_frames_sig}
    \end{subfigure}
    \caption{Summed frames within the fibre bundle region, separated based on whether they are selected or not with the Standard Method (Sec.~\ref{sec:standard_method}) and with the VAE (either BCE or KLD loss exceeded the 98th percentile on the $\mathrm{BG}$ test set).}
    \label{fig:sum_frames_98}
\end{figure}

\begin{figure}[ht]
    \centering
    \includegraphics[width=0.6\textwidth]{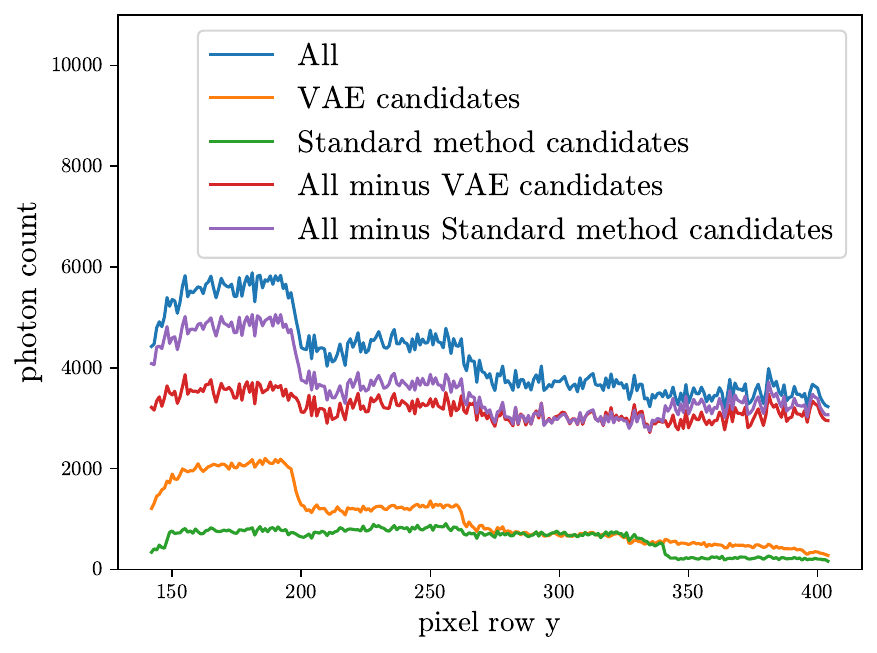}
    \caption{Photon count for each pixel row after summing all the frames for different scenarios on the $^{90}\mathrm{Sr+BG}$ sample.}
    \label{fig:comparison_methods}
\end{figure}

An example of our anomaly detection performance is presented in Fig.~\ref{fig:sum_frames_98}, which compares the standard method described in Sec.~\ref{sec:standard_method} with the VAE. Frames are selected in both the $\mathrm{BG}$ test set and the $^{90}\mathrm{Sr+BG}$ dataset. The VAE utilises the 98th percentile of the BCE and KLD losses from the $\mathrm{BG}$ test set to select frames. For the $\mathrm{BG}$ sample, the separation appears completely random to the naked eye for both cases (Fig.~\ref{fig:sum_frames_bkg}). On the other hand, for the $^{90}\mathrm{Sr+BG}$, the figure illustrates the VAE's ability to extract signal frames, where the pattern of short electron tracks traversing from top to bottom is distinctly visible, in contrast to the remaining candidates, where the summed frames resemble random noise (Fig.~\ref{fig:sum_frames_sig}). The selection made by the standard method seems more arbitrary, and clear signal patters seem to remain in the non-selected frames. This behaviour is also evident in Fig.~\ref{fig:comparison_methods}, where the total photon count after excluding the VAE-selected candidates exhibits a significantly flatter distribution compared to the result obtained by subtracting the standard method's track candidates, meaning that it rejects dark count events more efficiently. This observation indicates that the VAE is more effective at capturing signal events.

\begin{figure}[htbp]
    \centering
    \begin{subfigure}{0.49\textwidth}
        \centering
        \includegraphics[width=0.98\linewidth]{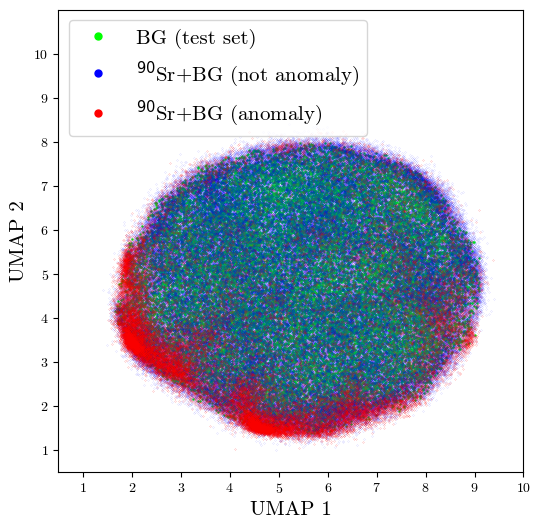}
        \caption{UMAP.}
        \label{fig:umap}
    \end{subfigure}
    \hfill
    \begin{subfigure}[b]{0.49\textwidth}
        \centering
        \includegraphics[width=1.\linewidth]{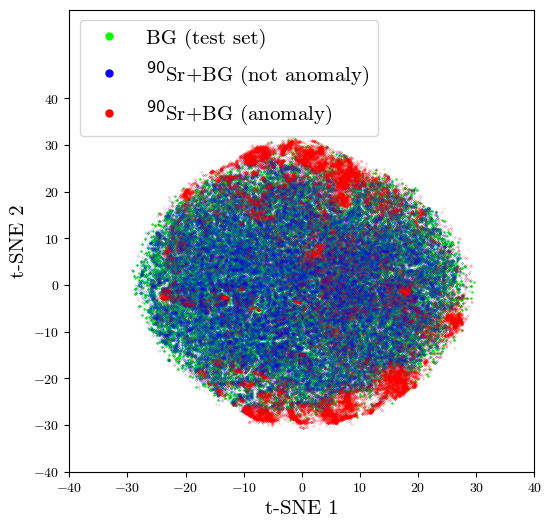}
        \caption{t-SNE.}
        \label{fig:tsne}
    \end{subfigure}
    \caption{2D projections of the 32-dimensional latent space for the $\mathrm{BG}$ test set and the $^{90}\mathrm{Sr+BG}$ dataset. The anomaly candidates are tagged following the selection from Fig.~\ref{fig:sum_frames_98}.}
    \label{fig:dim_red}
\end{figure}

To facilitate the detection of anomalies within the 32-dimensional latent space of the VAE, we employed two widely used dimensionality reduction techniques, Uniform Manifold Approximation and Projection (UMAP)~\cite{McInnes2018} and t-distributed Stochastic Neighbor Embedding (t-SNE)~\cite{JMLR:v9:vandermaaten08a}. These methods enable the projection of high-dimensional data into lower-dimensional spaces (2D in our case), allowing for the visualisation of latent space structures and identifying potential outliers (Fig.~\ref{fig:dim_red}); in both plots, most of the points are organised in a circular shape—likely due to the Gaussianity enforcement in the latent representation—reflecting the VAE's 32-dimensional latent space structure. The anomaly points are distinctly positioned along the borders of the the circles, deviating from the dense cluster of normal points. This spatial separation suggests that the latent space contains enough information to detect signal in the frames.

Table \ref{tab:my-table} shows the correlation between the number of counts and selected frames for both methods. For the VAE, the percentage of selected frames as anomaly candidates is higher in the $^{90}\mathrm{Sr+BG}$ sample compared to the $\mathrm{BG}$ test set for every number of counts per frame. In most cases, it is significantly higher or at least the same. We can conclude that the selection of a frame is not due to its higher number of counts but rather due to the signal event it likely contains. Of course, depending on the goal of the analysis, one can lower the threshold used in the BCE and KLD losses to select anomalies to avoid missing signals, but at the cost of allowing more background frames (high selection efficiency) or minimise the background at the risk of missing some signals (high selection purity). Additional results using different configurations of loss terms and thresholds are provided in Appendix~\ref{sec:appendix}. Both the KLD and BCE loss terms prove useful for anomaly detection. In contrast, the standard method, which was designed for frames across all count ranges (including those with more than ten counts), exhibits similar selection rates for both the $\mathrm{BG}$ test set and $^{90}\mathrm{Sr+BG}$ sample in the 5-9 counts per frame range. This highlights the challenge of anomaly detection in low-count scenarios, where traditional selection criteria fail to distinguish between background and signal frames effectively. Unlike the VAE-based approach, which adapts to subtle distribution shifts in the anomaly detection process, the standard method struggles to capture meaningful differences, leading to a more uniform selection rate across both samples.

Although standard anomaly detection metrics such as precision, recall, or AUC are commonly used to evaluate classification performance, their application in this study is limited by the absence of ground truth labels. Unlike simulated datasets where the true classification of each frame is known, the experimental nature of our dataset does not allow a precise quantification of detection performance. However, the results presented in this section demonstrate that our method successfully isolates signal events, as summing anomalous and non-anomalous frames separately reveals a clear track pattern for the detected anomalies.

\begin{table}[htbp]
    \centering
    \renewcommand{\arraystretch}{1.1} 
    \setlength{\tabcolsep}{6pt} 
    \resizebox{0.9\textwidth}{!}{ 
    \begin{tabular}{c|r|>{\columncolor{gray!20}}r>{\columncolor{gray!20}}r|r r|r|>{\columncolor{gray!20}}r>{\columncolor{gray!20}}r|r r}
        \toprule
        \multirow{3}{*}{\textbf{Counts}} & \multicolumn{5}{c|}{\cellcolor{gray!30}\textbf{BG (test)}} & \multicolumn{5}{c}{\textbf{$^{90}\mathrm{Sr+BG}$}} \\ \cline{2-11}
        & \multirow{2}{*}{\textbf{Frames}} & \multicolumn{2}{c|}{\textbf{VAE}} & \multicolumn{2}{c|}{\textbf{Standard}} & \multirow{2}{*}{\textbf{Frames}} & \multicolumn{2}{c|}{\textbf{VAE}} & \multicolumn{2}{c}{\textbf{Standard}} \\ \cline{3-6} \cline{8-11} 
        &  & \textbf{Selected} & \textbf{\%} & \textbf{Selected} & \textbf{\%} &  & \textbf{Selected} & \textbf{\%} & \textbf{Selected} & \textbf{\%} \\ 
        \midrule
        4 & 23396 & 104 & 0.44 & 1865 & 7.97 & 153066 & 2944 & 1.92 & 12400 & 8.10 \\
        \rowcolor{gray!10} 5 & 4830 & 367 & 7.60 & 813 & 16.83 & 44146 & 13630 & 30.87 & 6903 & 15.64 \\
        6 & 828 & 424 & 51.21 & 220 & 26.57 & 14397 & 12457 & 86.52 & 3020 & 20.98 \\
        \rowcolor{gray!10} 7 & 109 & 95 & 87.16 & 46 & 42.20 & 7521 & 7494 & 99.64 & 1710 & 22.74 \\
        8 & 14 & 14 & 100.00 & 4 & 28.57 & 5768 & 5768 & 100.00 & 1469 & 25.47 \\
        \rowcolor{gray!10} 9 & 1 & 1 & 100.00 & 1 & 100.00 & 5159 & 5159 & 100.00 & 1618 & 31.36 \\
        \bottomrule
    \end{tabular}
    }
    \caption{Comparison of the fraction of selected frames for the $\mathrm{BG}$ test set and the \textbf{$^{90}\mathrm{Sr+BG}$} sample across different numbers of counts per frame. Frames are identified as anomalies if either the BCE or KLD loss exceeds the 98th percentile on the $\mathrm{BG}$ test set.}
    \label{tab:my-table}
\end{table}

\subsection{Fast inference}
\label{sec:fast_inference}

The computational performance of the VAE for anomaly detection was evaluated on both CPU and GPU platforms, with the results summarised in Table~\ref{tab:time_cpu_gpu}. The computation times, reported in milliseconds per frame, illustrate the VAE's efficiency in processing batches of sensor data. Both CPU and GPU platforms significantly reduce processing time per frame as batch size increases. However, the GPU consistently outperforms the CPU, achieving computation times up to two orders of magnitude lower, with the fastest time recorded at just 0.08 ms per frame for a batch size of 64. These findings underscore the VAE's potential for real-time signal detection tasks, particularly when utilising GPU hardware, which provides substantial speedups over CPU processing.

\begin{table}[htbp]
    \centering
    \renewcommand{\arraystretch}{1.2} 
    \setlength{\tabcolsep}{10pt} 
    \resizebox{1.0\textwidth}{!}{ 
    \begin{tabular}{c|c|>{\columncolor{gray!20}}r>{\columncolor{gray!20}}r>{\columncolor{gray!20}}r>{\columncolor{gray!20}}r>{\columncolor{gray!20}}r>{\columncolor{gray!20}}r>{\columncolor{gray!20}}r}
        \toprule
        \multirow{3}{*}{\textbf{Phase}} & \multirow{3}{*}{\textbf{Device}} & \multicolumn{7}{c}{\cellcolor{gray!30}\textbf{Batch size (number of frames processed in parallel)}} \\ \cline{3-9} 
        & & \textbf{1} & \textbf{2} & \textbf{4} & \textbf{8} & \textbf{16} & \textbf{32} & \textbf{64} \\ 
        & & \multicolumn{7}{c}{\cellcolor{gray!10}\textbf{Time per frame [ms]}} \\ 
        \midrule
        \multirow{2}{*}{\textbf{Encoder}} & \textbf{CPU} & 3.7 $\pm$ 1.2 & 1.8 $\pm$ 0.6 & 1.3 $\pm$ 0.7 & 0.7 $\pm$ 0.4 & 0.4 $\pm$ 0.2 & 0.2 $\pm$ 0.1 & 0.1 $\pm$ 0.1 \\
        & {GPU} & \cellcolor{gray!10} 1.6 $\pm$ 7.5  & \cellcolor{gray!10} 0.7 $\pm$ 0.9  & \cellcolor{gray!10} 0.4 $\pm$ 0.1  & \cellcolor{gray!10} 0.2 $\pm$ 0.2  & \cellcolor{gray!10} 0.1 $\pm$ 0.1  & \cellcolor{gray!10} 0.1 $\pm$ 0.1  & \cellcolor{gray!10} 0.0 $\pm$ 0.0 \\
        \midrule
        \multirow{2}{*}{\textbf{Decoder}} & \textbf{CPU} & 19.1 $\pm$ 1.9 & 15.7 $\pm$ 0.7 & 15.4 $\pm$ 0.7 & 15.4 $\pm$ 0.5 & 15.1 $\pm$ 0.5 & 14.0 $\pm$ 4.3 & 7.0 $\pm$ 7.6 \\
        & {GPU} & \cellcolor{gray!10} 0.5 $\pm$ 7.0  & \cellcolor{gray!10} 0.3 $\pm$ 0.9  & \cellcolor{gray!10} 0.1 $\pm$ 0.1  & \cellcolor{gray!10} 0.1 $\pm$ 0.1  & \cellcolor{gray!10} 0.1 $\pm$ 0.1  & \cellcolor{gray!10} 0.1 $\pm$ 0.1  & \cellcolor{gray!10} 0.0 $\pm$ 0.1 \\
        \midrule
        \multirow{2}{*}{\textbf{KLD}} & \textbf{CPU} & 0.2 $\pm$ 0.0 & 0.1 $\pm$ 0.0 & 0.1 $\pm$ 0.0 & 0.0 $\pm$ 0.0 & 0.0 $\pm$ 0.0 & 0.0 $\pm$ 0.0 & 0.0 $\pm$ 0.0 \\
        & {GPU} & \cellcolor{gray!10} 0.3 $\pm$ 1.4  & \cellcolor{gray!10} 0.1 $\pm$ 0.0  & \cellcolor{gray!10} 0.1 $\pm$ 0.0  & \cellcolor{gray!10} 0.0 $\pm$ 0.0  & \cellcolor{gray!10} 0.0 $\pm$ 0.0  & \cellcolor{gray!10} 0.0 $\pm$ 0.0  & \cellcolor{gray!10} 0.0 $\pm$ 0.0 \\
        \midrule
        \multirow{2}{*}{\textbf{BCE}} & \textbf{CPU} & 0.9 $\pm$ 0.3 & 0.5 $\pm$ 0.2 & 0.3 $\pm$ 0.1 & 0.2 $\pm$ 0.0 & 0.1 $\pm$ 0.0 & 0.1 $\pm$ 0.0 & 0.0 $\pm$ 0.0 \\
        & {GPU} & \cellcolor{gray!10} 0.1 $\pm$ 0.5  & \cellcolor{gray!10} 0.1 $\pm$ 0.0  & \cellcolor{gray!10} 0.0 $\pm$ 0.0  & \cellcolor{gray!10} 0.0 $\pm$ 0.0  & \cellcolor{gray!10} 0.0 $\pm$ 0.0  & \cellcolor{gray!10} 0.0 $\pm$ 0.0  & \cellcolor{gray!10} 0.0 $\pm$ 0.0 \\
        \midrule
        \multirow{1}{*}{\textbf{Memory allocation}} & {GPU} & \cellcolor{gray!10} 0.2 $\pm$ 0.0 & \cellcolor{gray!10} 0.1 $\pm$ 0.0 & \cellcolor{gray!10} 0.1 $\pm$ 0.0 & \cellcolor{gray!10} 0.1 $\pm$ 0.0 & \cellcolor{gray!10} 0.1 $\pm$ 0.0 & \cellcolor{gray!10} 0.0 $\pm$ 0.0 & \cellcolor{gray!10} 0.0 $\pm$ 0.0 \\
        \midrule
        \multirow{2}{*}{\textbf{Total}} & \textbf{CPU} & 23.8 $\pm$ 1.5 & 18.2 $\pm$ 0.7 & 17.1 $\pm$ 0.6 & 16.2 $\pm$ 0.6 & 15.5 $\pm$ 0.5 & 14.2 $\pm$ 4.4 & 7.1 $\pm$ 7.7 \\
        & {GPU} & \cellcolor{gray!10} 2.6 $\pm$ 2.5  & \cellcolor{gray!10} 1.3 $\pm$ 0.1  & \cellcolor{gray!10} 0.7 $\pm$ 0.1  & \cellcolor{gray!10} 0.4 $\pm$ 0.0  & \cellcolor{gray!10} 0.3 $\pm$ 0.0  & \cellcolor{gray!10} 0.2 $\pm$ 0.1  & \cellcolor{gray!10} 0.1 $\pm$ 0.1 \\
        \bottomrule
    \end{tabular}
    }
    \caption{Comparison of computation times for different batch sizes on CPU (Intel(R) Xeon(R) CPU E5-2698 v4 @ 2.20GHz) and GPU (Tesla V100-SXM2) systems. The time reported is the VAE processing time per frame, and each result is an average of 1000 independent computations. The system has a total memory of 528 GB.}
    \label{tab:time_cpu_gpu}
\end{table}

\section{Conclusion}
\label{sec:conclusion}

Scintillating fibres read out with imaging sensors offer an optimal configuration for high spatial resolution particle tracking. Given the growing success of machine learning techniques for anomaly detection, we proposed using a VAE to differentiate between signal and background frames. In this study, we demonstrated that a VAE, trained solely on background frames, can serve as an effective and fast selection algorithm. We used the BCE and the KLD losses as anomaly scores to identify signal frames. The results showed that the frames selected indeed contained signal events. Furthermore, we tested the same method on background frames and found that significantly fewer frames were selected, confirming the effectiveness of the VAE with both BCE and KLD losses as anomaly detection metrics. In addition, a comparison with a standard selection method demonstrated that the VAE approach significantly outperforms traditional techniques, particularly in challenging low-count scenarios. Although we have not yet tested this approach with finer fibres, we hypothesise that the VAE would further improve its performance, given the lower number of dark counts and the more refined pattern left by a particle, which would be easier and faster for the method to detect. In future work, we plan to explore its integration as a real-time anomaly detection method into the hardware.

\appendix
\section{Additional results}
\label{sec:appendix}

This appendix provides supplementary results, demonstrating the impact of varying thresholds on BCE and KLD loss values of the VAE. The results highlight how using these loss functions individually or in combination yields different sets of anomalies. Specifically, the following sections focus on anomaly selection based on the criteria detailed below:

\begin{itemize}
    \item Section~\ref{sec:bce}: Selection of anomalies using thresholds applied solely to the BCE loss.
    \item Section~\ref{sec:kld}: Selection of anomalies using thresholds applied solely to the KLD loss.
    \item Section~\ref{sec:both}: Selection of anomalies using thresholds applied to both the BCE and KLD losses.
\end{itemize}

In each section, the following thresholds are applied to the respective loss functions to identify anomaly candidates, as discussed in Section ~\ref{sec:anomaly_detection_vae}:

\begin{itemize}
    \item \textbf{Divergence threshold:} The point at which the $^{90}\mathrm{Sr+BG}$ loss distributions start to exceed those of the $\mathrm{BG}$ test set, as illustrated in Fig.~\ref{fig:losses_comparison}.
    \item \textbf{98th percentile threshold:} The 98th percentile of the loss distribution for the $\mathrm{BG}$ test set.
    \item \textbf{Max threshold:} The maximum loss value observed in the $\mathrm{BG}$ test set.
\end{itemize}

\subsection{Anomaly detection on BCE}
\label{sec:bce}

\begin{figure}[htbp]
    \centering

    \begin{subfigure}{\textwidth}
        \centering
        \begin{subfigure}{0.45\textwidth}
            \centering
            \includegraphics[width=1.\linewidth]{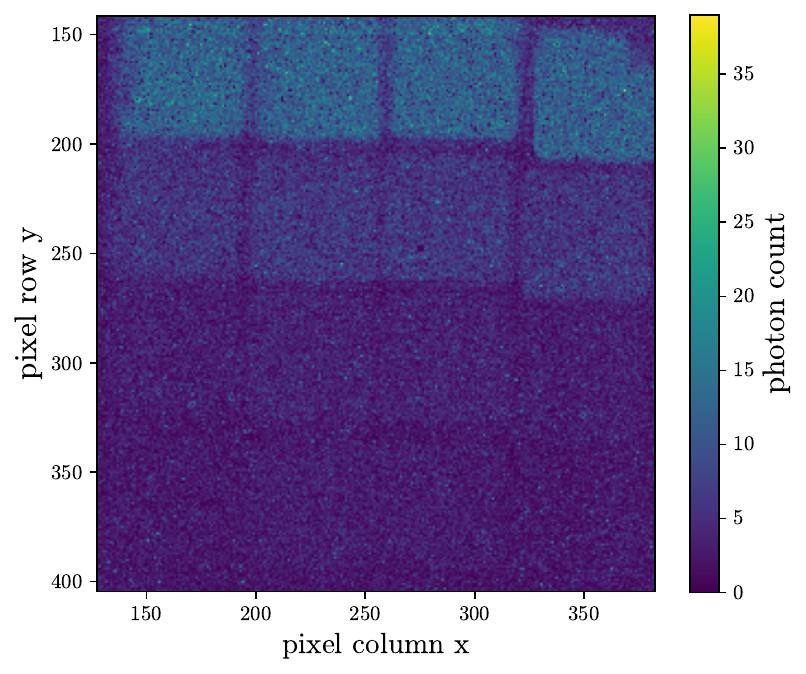}
            \caption*{Anomaly candidate frames summed.}
            \label{fig:bce_ano1}
        \end{subfigure}
        \hfill
        \begin{subfigure}{0.45\textwidth}
            \centering
            \includegraphics[width=1.\linewidth]{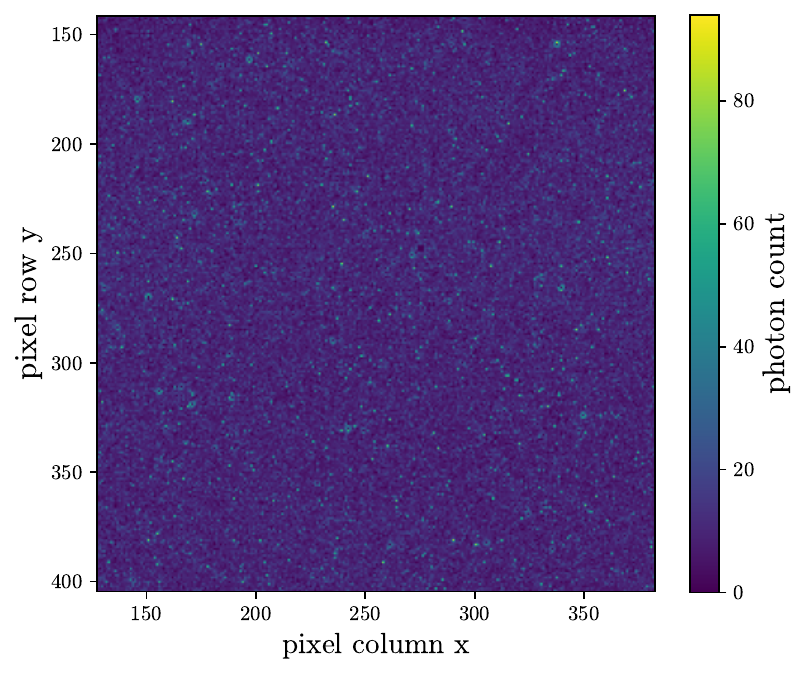}
            \caption*{Non-anomaly candidate frames summed.}
            \label{fig:bce_nor1}
        \end{subfigure}
        \caption*{(a) Divergence threshold ($^{90}\mathrm{Sr+BG}$ > $\mathrm{BG}$ test set).}
        \label{fig:bce_row1}
    \end{subfigure}

    \vspace{1em}

    \begin{subfigure}{\textwidth}
        \centering
        \begin{subfigure}{0.45\textwidth}
            \centering
            \includegraphics[width=1.\linewidth]{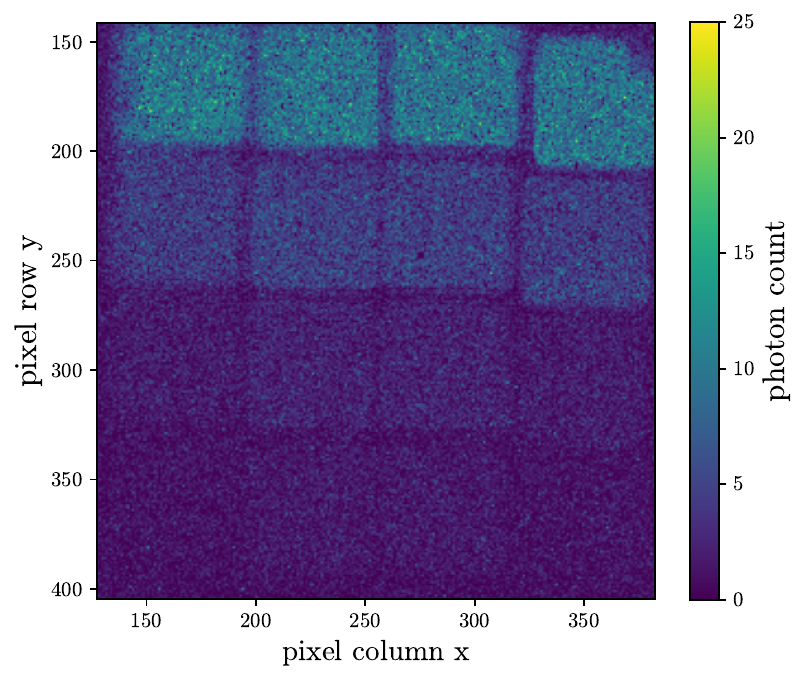}
            \caption*{Anomaly candidate frames summed.}
            \label{fig:bce_ano2}
        \end{subfigure}
        \hfill
        \begin{subfigure}{0.45\textwidth}
            \centering
            \includegraphics[width=1.\linewidth]{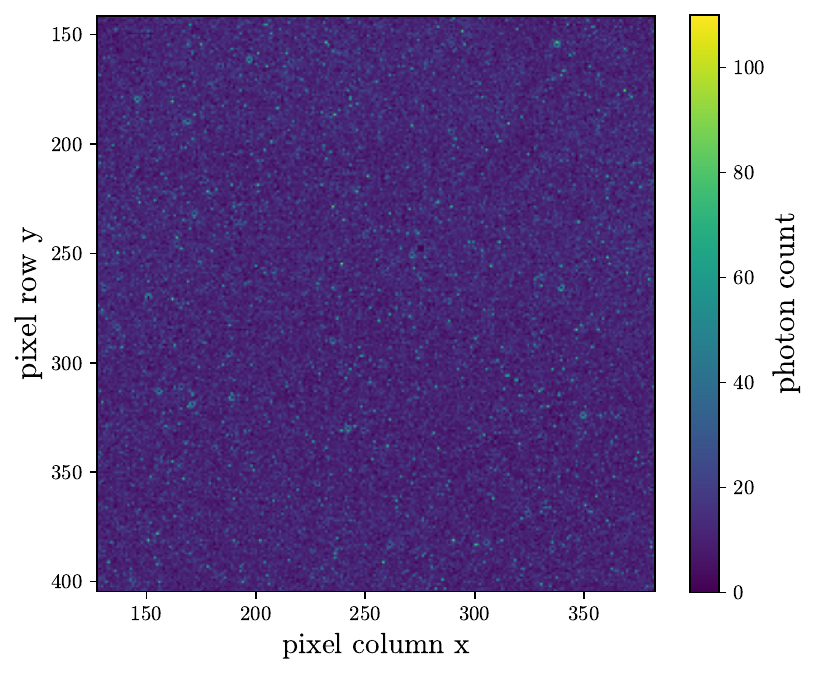}
            \caption*{Non-anomaly candidate frames summed.}
            \label{fig:bce_nor2}
        \end{subfigure}
        \caption*{(b) 98th percentile threshold ($\mathrm{BG}$ test set).}
        \label{fig:bce_row2}
    \end{subfigure}

    \vspace{1em} 

    \begin{subfigure}{\textwidth}
        \centering
        \begin{subfigure}{0.45\textwidth}
            \centering
            \includegraphics[width=1.\linewidth]{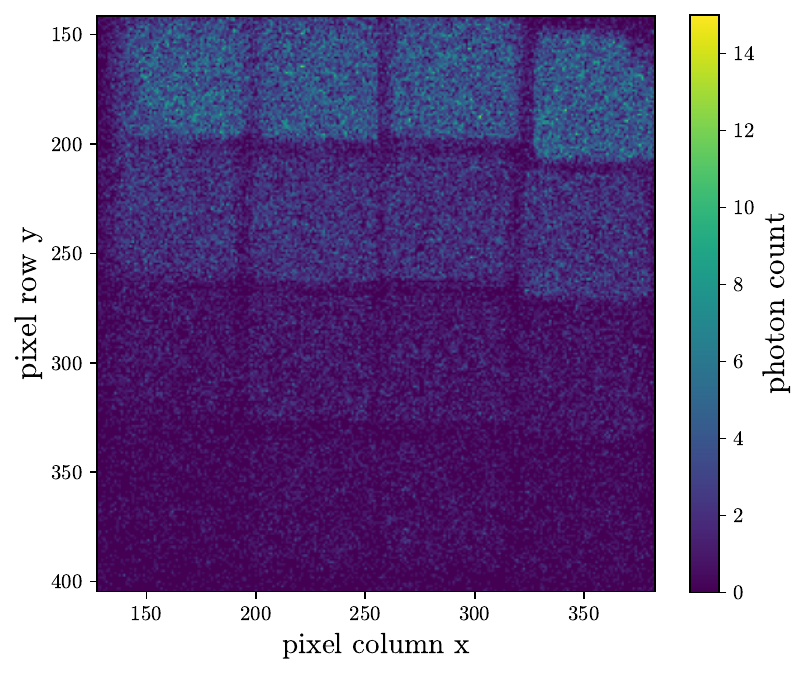}
            \caption*{Anomaly candidate frames summed.}
            \label{fig:bce_ano3}
        \end{subfigure}
        \hfill
        \begin{subfigure}{0.45\textwidth}
            \centering
            \includegraphics[width=1.\linewidth]{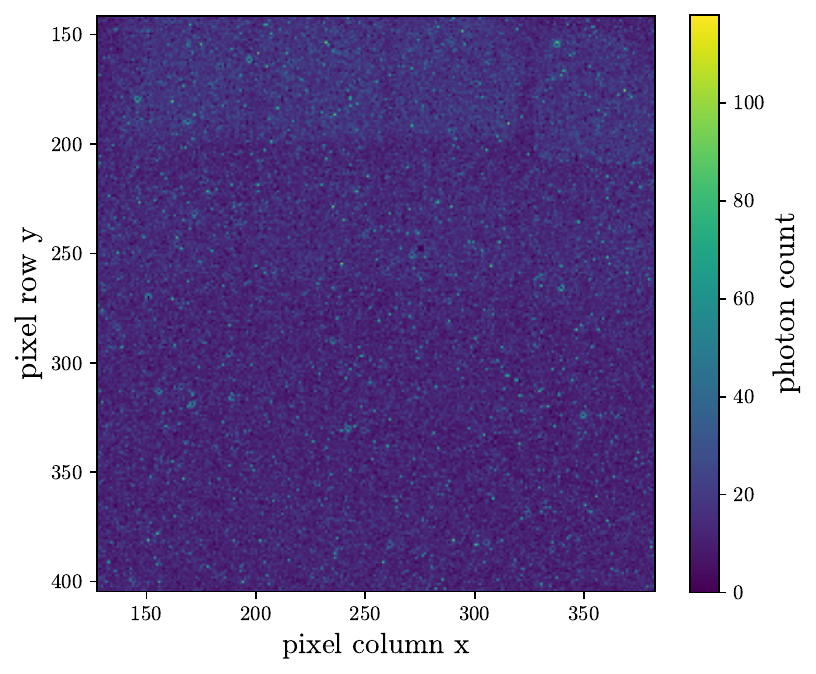}
            \caption*{Non-anomaly candidate frames summed.}
            \label{fig:bce_nor3}
        \end{subfigure}
        \caption*{(c) Max threshold ($\mathrm{BG}$ test set).}
        \label{fig:bce_row3}
    \end{subfigure}

    \caption{Summed frames within the fibre bundle region for the $^{90}\mathrm{Sr+BG}$ dataset, separated based on whether they are considered anomalies using the BCE loss.}
    \label{fig:bce}
\end{figure}

Figure~\ref{fig:bce} shows the total sum of frames selected and not selected as anomaly candidates based on different thresholds applied to the BCE loss values, while Table~\ref{tab:bce} presents the total number of frames selected in each case, categorised by the number of photon counts in the frame. This example clearly illustrates that applying a higher threshold to the BCE loss value results in fewer background frames being selected as signal candidates, while more genuine signal candidates are overlooked; and vice versa.

\begin{table}[htbp]
    \centering
    \renewcommand{\arraystretch}{1.1} 
    \setlength{\tabcolsep}{8pt} 
    \resizebox{0.7\textwidth}{!}{ 
    \begin{tabular}{c|>{\columncolor{gray!20}}r|>{\columncolor{gray!20}}r|>{\columncolor{gray!20}}r|r|r|r}
        \toprule
        \multicolumn{7}{c}{\textbf{Divergence threshold ($^{90}\mathrm{Sr+BG}$ > $\mathrm{BG}$ test set)}} \\ \midrule
        \multirow{2}{*}{\textbf{Counts}} & \multicolumn{3}{c|}{\cellcolor{gray!30}\textbf{BG (test)}} & \multicolumn{3}{c}{\textbf{$^{90}\mathrm{Sr+BG}$}} \\ \cline{2-7} 
         & \textbf{Frames} & \textbf{Selected} & \textbf{\%} & \textbf{Frames} & \textbf{Selected} & \textbf{\%} \\ 
        \midrule
        4 & 23396 & 1243 & 5.31 & 153066 & 16408 & 10.72 \\
        \rowcolor{gray!10} 5 & 4830 & 1178 & 24.39 & 44146 & 18459 & 41.81 \\
        6 & 828 & 520 & 62.80 & 14397 & 12134 & 84.28 \\
        \rowcolor{gray!10} 7 & 109 & 100 & 91.74 & 7521 & 7462 & 99.22 \\
        8 & 14 & 14 & 100.00 & 5768 & 5766 & 99.97 \\
        \rowcolor{gray!10} 9 & 1 & 1 & 100.00 & 5159 & 5159 & 100.00 \\
        \bottomrule
        \toprule
        \multicolumn{7}{c}{\textbf{98th Percentile threshold ($\mathrm{BG}$ test set)}} \\ \midrule
        \multirow{2}{*}{\textbf{Counts}} & \multicolumn{3}{c|}{\cellcolor{gray!30}\textbf{BG (test)}} & \multicolumn{3}{c}{\textbf{$^{90}\mathrm{Sr+BG}$}} \\ \cline{2-7} 
         & \textbf{Frames} & \textbf{Selected} & \textbf{\%} & \textbf{Frames} & \textbf{Selected} & \textbf{\%} \\ 
        \midrule
        4 & 23396 & 100 & 0.43 & 153066 & 2632 & 1.72 \\
        \rowcolor{gray!10} 5 & 4830 & 207 & 4.29 & 44146 & 6828 & 15.47 \\
        6 & 828 & 197 & 23.79 & 14397 & 8448 & 58.68 \\
        \rowcolor{gray!10} 7 & 109 & 66 & 60.55 & 7521 & 7114 & 94.59 \\
        8 & 14 & 13 & 92.86 & 5768 & 5762 & 99.90 \\
        \rowcolor{gray!10} 9 & 1 & 1 & 100.00 & 5159 & 5159 & 100.00 \\
        \bottomrule
        \toprule
        \multicolumn{7}{c}{\textbf{Max threshold ($\mathrm{BG}$ test set)}} \\ \midrule
        \multirow{2}{*}{\textbf{Counts}} & \multicolumn{3}{c|}{\cellcolor{gray!30}\textbf{BG (test)}} & \multicolumn{3}{c}{\textbf{$^{90}\mathrm{Sr+BG}$}} \\ \cline{2-7} 
         & \textbf{Frames} & \textbf{Selected} & \textbf{\%} & \textbf{Frames} & \textbf{Selected} & \textbf{\%} \\ 
        \midrule
        4 & 23396 & 0 & 0.00 & 153066 & 4 & 0.00 \\
        \rowcolor{gray!10} 5 & 4830 & 0 & 0.00 & 44146 & 8 & 0.02 \\
        6 & 828 & 0 & 0.00 & 14397 & 177 & 1.23 \\
        \rowcolor{gray!10} 7 & 109 & 0 & 0.00 & 7521 & 1350 & 17.95 \\
        8 & 14 & 0 & 0.00 & 5768 & 3490 & 60.51 \\
        \rowcolor{gray!10} 9 & 1 & 0 & 0.00 & 5159 & 4750 & 92.07 \\
        \bottomrule
    \end{tabular}
    }
    
    \caption{Comparison of the fraction of selected frames using the BCE loss for the $\mathrm{BG}$ testing set and the \textbf{$^{90}\mathrm{Sr+BG}$} sample for different numbers of counts per frame.}
    \label{tab:bce}
\end{table}

\subsection{Anomaly detection on KLD}
\label{sec:kld}

\begin{figure}[htbp]
    \centering

    \begin{subfigure}{\textwidth}
        \centering
        \begin{subfigure}{0.45\textwidth}
            \centering
            \includegraphics[width=1.\linewidth]{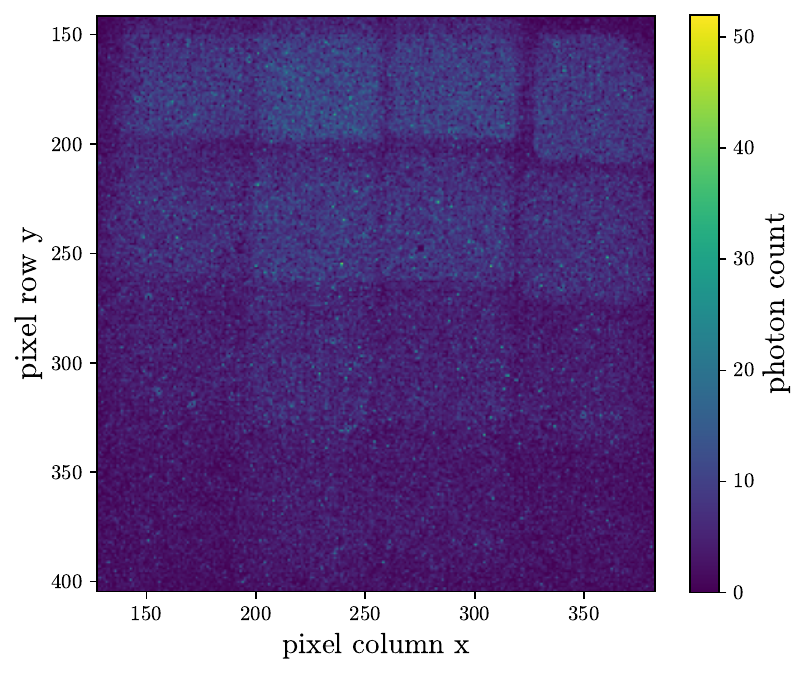}
            \caption*{Anomaly candidate frames summed.}
            \label{fig:kld_ano1}
        \end{subfigure}
        \hfill
        \begin{subfigure}{0.45\textwidth}
            \centering
            \includegraphics[width=1.\linewidth]{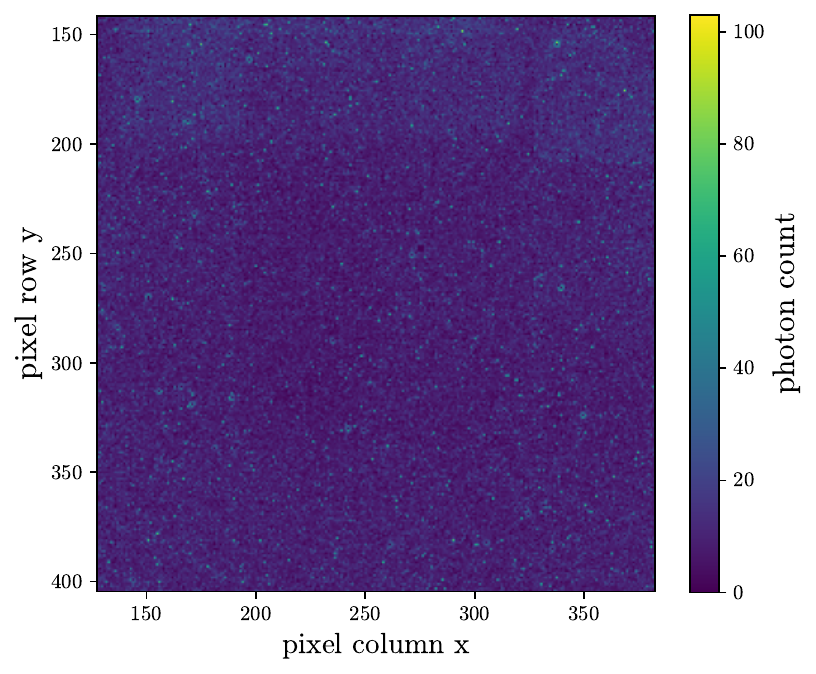}
            \caption*{Non-anomaly candidate frames summed.}
            \label{fig:kld_nor1}
        \end{subfigure}
        \caption*{(a) Divergence threshold ($^{90}\mathrm{Sr+BG}$ > $\mathrm{BG}$ test set).}
        \label{fig:kld_row1}
    \end{subfigure}

    \vspace{1em} 

    \begin{subfigure}{\textwidth}
        \centering
        \begin{subfigure}{0.45\textwidth}
            \centering
            \includegraphics[width=1.\linewidth]{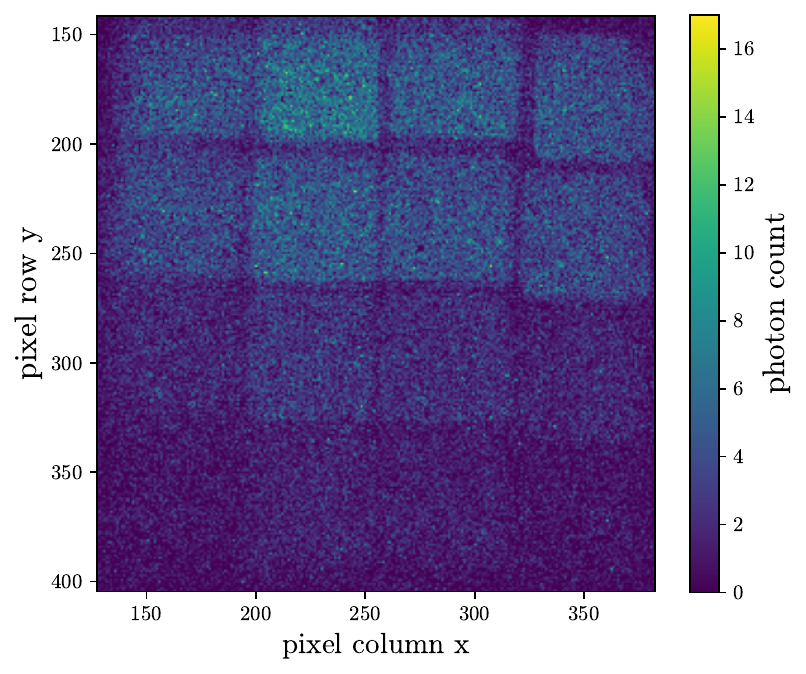}
            \caption*{Anomaly candidate frames summed.}
            \label{fig:kld_ano2}
        \end{subfigure}
        \hfill
        \begin{subfigure}{0.45\textwidth}
            \centering
            \includegraphics[width=1.\linewidth]{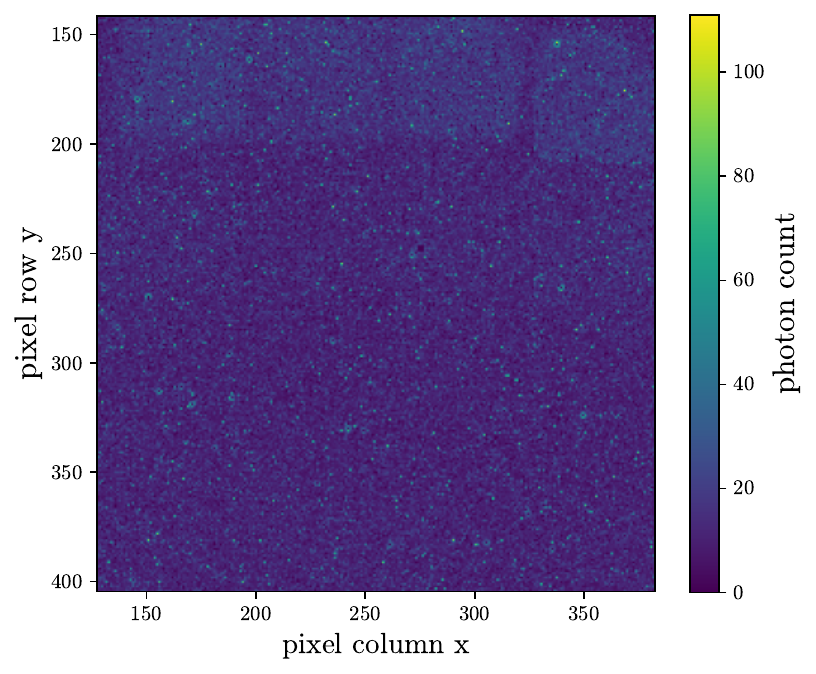}
            \caption*{Non-anomaly candidate frames summed.}
            \label{fig:kld_nor2}
        \end{subfigure}
        \caption*{(b) 98th Percentile threshold ($\mathrm{BG}$ test set).}
        \label{fig:kld_row2}
    \end{subfigure}

    \vspace{1em} 

    \begin{subfigure}{\textwidth}
        \centering
        \begin{subfigure}{0.45\textwidth}
            \centering
            \includegraphics[width=1.\linewidth]{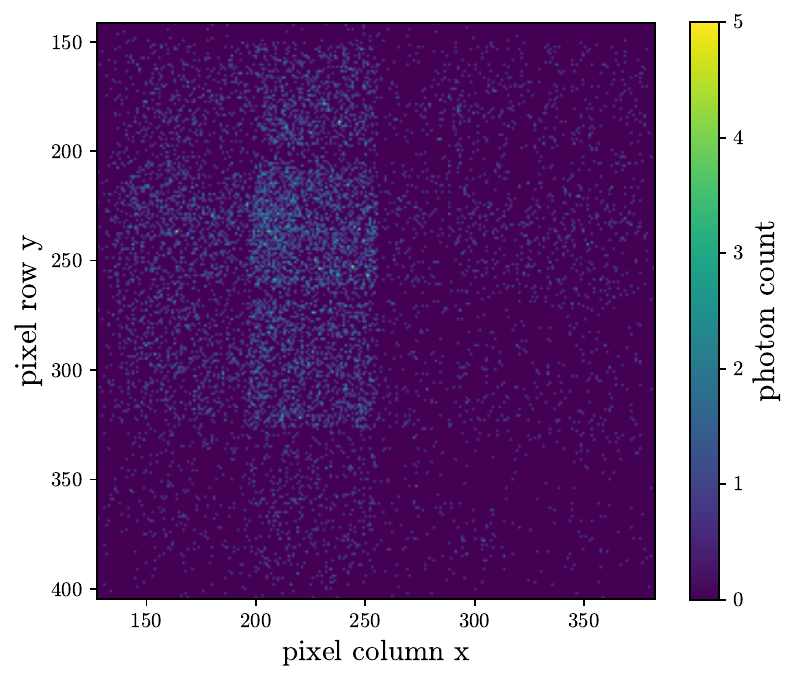}
            \caption*{Anomaly candidate frames summed.}
            \label{fig:kld_ano3}
        \end{subfigure}
        \hfill
        \begin{subfigure}{0.45\textwidth}
            \centering
            \includegraphics[width=1.\linewidth]{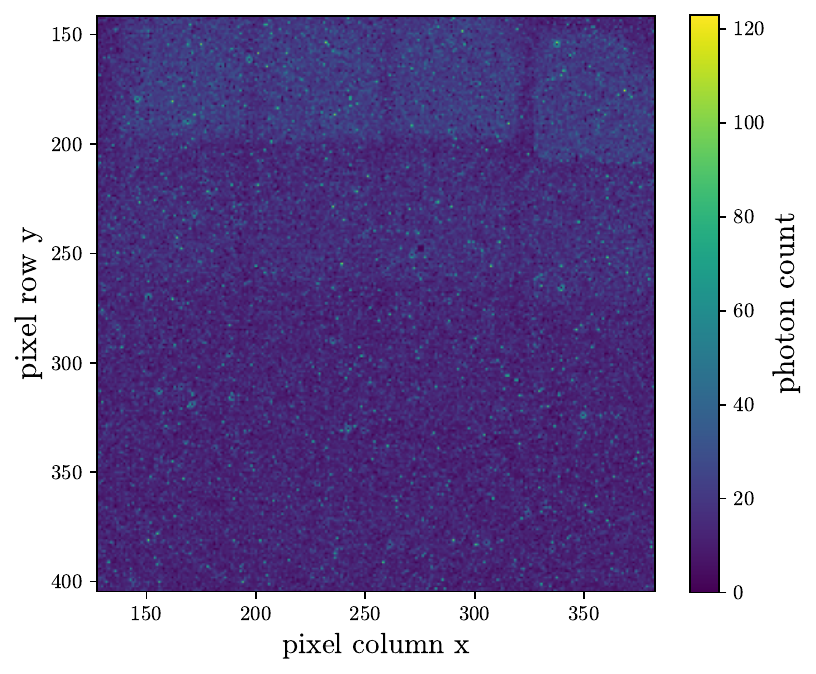}
            \caption*{Non-anomaly candidate frames summed.}
            \label{fig:kld_nor3}
        \end{subfigure}
        \caption*{(c) Max threshold ($\mathrm{BG}$ test set).}
        \label{fig:kld_row3}
    \end{subfigure}

    \caption{Summed frames within the fibre bundle region for the $^{90}\mathrm{Sr+BG}$ dataset, separated based on whether they are considered anomalies using the KLD loss.}
    \label{fig:kld}
\end{figure}

Figure~\ref{fig:kld} shows the total sum of frames selected and not selected as anomaly candidates based on different thresholds applied to the KLD loss values, while Table~\ref{tab:kld} presents the total number of frames selected in each case, categorised by the number of photon counts in the frame. The visual separation between signal and background frames suggests that the KLD loss may not be as effective as the BCE loss for this particular case study. However, as shown in Fig.~\ref{fig:dim_red}, the latent space clearly encodes signal events differently, indicating that further investigation would be required. This would be especially important if one aims to reduce the computational complexity of the VAE by running only the encoder and detecting anomalies based solely on the latent representation.

\begin{table}[htbp]
    \centering
    \renewcommand{\arraystretch}{1.1} 
    \setlength{\tabcolsep}{8pt} 
    \resizebox{0.7\textwidth}{!}{ 
    \begin{tabular}{c|>{\columncolor{gray!20}}r|>{\columncolor{gray!20}}r|>{\columncolor{gray!20}}r|r|r|r}
        \toprule
        \multicolumn{7}{c}{\textbf{Divergence threshold ($^{90}\mathrm{Sr+BG}$ > $\mathrm{BG}$ test set)}} \\ \midrule
        \multirow{2}{*}{\textbf{Counts}} & \multicolumn{3}{c|}{\cellcolor{gray!30}\textbf{BG (test)}} & \multicolumn{3}{c}{\textbf{$^{90}\mathrm{Sr+BG}$}} \\ \cline{2-7} 
         & \textbf{Frames} & \textbf{Selected} & \textbf{\%} & \textbf{Frames} & \textbf{Selected} & \textbf{\%} \\ 
        \midrule
        4 & 23396 & 1059 & 4.53 & 153066 & 33327 & 21.77 \\
        \rowcolor{gray!10} 5 & 4830 & 2606 & 53.95 & 44146 & 31695 & 71.80 \\
        6 & 828 & 754 & 91.06 & 14397 & 12324 & 85.60 \\
        \rowcolor{gray!10} 7 & 109 & 106 & 97.25 & 7521 & 6868 & 91.32 \\
        8 & 14 & 14 & 100.00 & 5768 & 5529 & 95.86 \\
        \rowcolor{gray!10} 9 & 1 & 1 & 100.00 & 5159 & 5086 & 98.58 \\
        \bottomrule
        \toprule
        \multicolumn{7}{c}{\textbf{98th Percentile threshold ($\mathrm{BG}$ test set)}} \\ \midrule
        \multirow{2}{*}{\textbf{Counts}} & \multicolumn{3}{c|}{\cellcolor{gray!30}\textbf{BG (test)}} & \multicolumn{3}{c}{\textbf{$^{90}\mathrm{Sr+BG}$}} \\ \cline{2-7} 
         & \textbf{Frames} & \textbf{Selected} & \textbf{\%} & \textbf{Frames} & \textbf{Selected} & \textbf{\%} \\ 
        \midrule
        4 & 23396 & 4 & 0.02 & 153066 & 321 & 0.21 \\
        \rowcolor{gray!10} 5 & 4830 & 174 & 3.60 & 44146 & 7604 & 17.22 \\
        6 & 828 & 309 & 37.32 & 14397 & 7691 & 53.42 \\
        \rowcolor{gray!10} 7 & 109 & 82 & 75.23 & 7521 & 5155 & 68.54 \\
        8 & 14 & 14 & 100.00 & 5768 & 4721 & 81.85 \\
        \rowcolor{gray!10} 9 & 1 & 1 & 100.00 & 5159 & 4746 & 91.99 \\
        \bottomrule
        \toprule
        \multicolumn{7}{c}{\textbf{Max threshold ($\mathrm{BG}$ test set)}} \\ \midrule
        \multirow{2}{*}{\textbf{Counts}} & \multicolumn{3}{c|}{\cellcolor{gray!30}\textbf{BG (test)}} & \multicolumn{3}{c}{\textbf{$^{90}\mathrm{Sr+BG}$}} \\ \cline{2-7} 
         & \textbf{Frames} & \textbf{Selected} & \textbf{\%} & \textbf{Frames} & \textbf{Selected} & \textbf{\%} \\ 
        \midrule
        4 & 23396 & 0 & 0.00 & 153066 & 0 & 0.00 \\
        \rowcolor{gray!10} 5 & 4830 & 0 & 0.00 & 44146 & 0 & 0.00 \\
        6 & 828 & 0 & 0.00 & 14397 & 0 & 0.00 \\
        \rowcolor{gray!10} 7 & 109 & 0 & 0.00 & 7521 & 18 & 0.24 \\
        8 & 14 & 0 & 0.00 & 5768 & 239 & 4.14 \\
        \rowcolor{gray!10} 9 & 1 & 0 & 0.00 & 5159 & 893 & 17.31 \\
        \bottomrule
    \end{tabular}
    }
    
    \caption{Comparison of the fraction of selected frames using the KLD loss for the $\mathrm{BG}$ testing set and the \textbf{$^{90}\mathrm{Sr+BG}$} sample for different numbers of counts per frame.}
    \label{tab:kld}
\end{table}

\subsection{Anomaly on both BCE and KLD}
\label{sec:both}

\begin{figure}[htbp]
    \centering

    \begin{subfigure}{\textwidth}
        \centering
        \begin{subfigure}{0.45\textwidth}
            \centering
            \includegraphics[width=1.\linewidth]{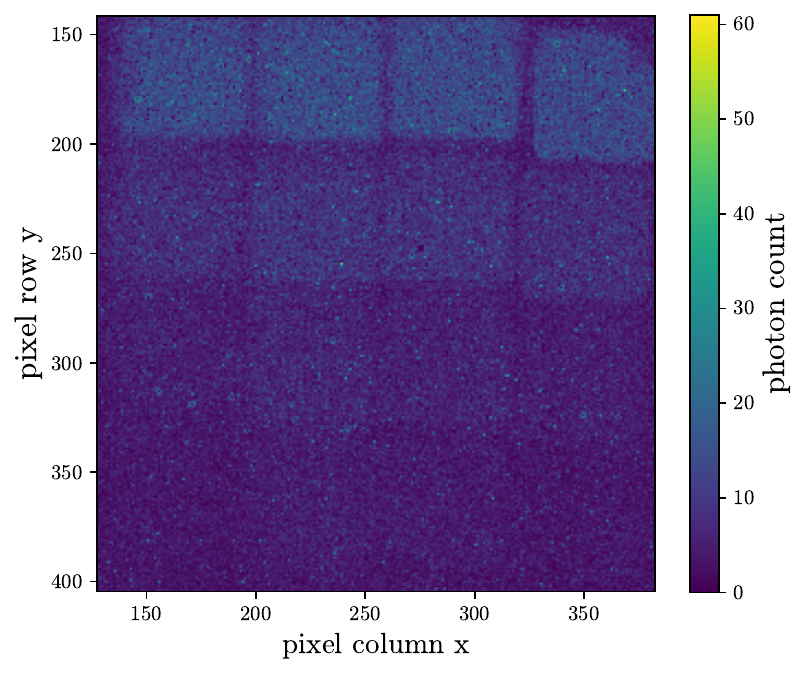}
            \caption*{Anomaly candidate frames summed.}
            \label{fig:both_ano1}
        \end{subfigure}
        \hfill
        \begin{subfigure}{0.45\textwidth}
            \centering
            \includegraphics[width=1.\linewidth]{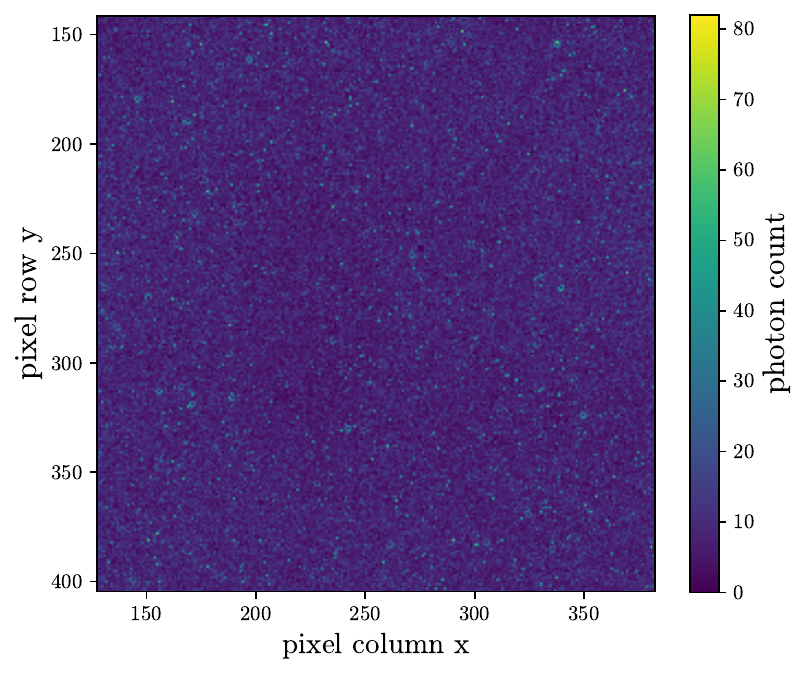}
            \caption*{Non-anomaly candidate frames summed.}
            \label{fig:both_nor1}
        \end{subfigure}
        \caption*{(a) Divergence threshold ($^{90}\mathrm{Sr+BG}$ > $\mathrm{BG}$ test set).}
        \label{fig:both_row1}
    \end{subfigure}

    \vspace{1em} 

    \begin{subfigure}{\textwidth}
        \centering
        \begin{subfigure}{0.45\textwidth}
            \centering
            \includegraphics[width=1.\linewidth]{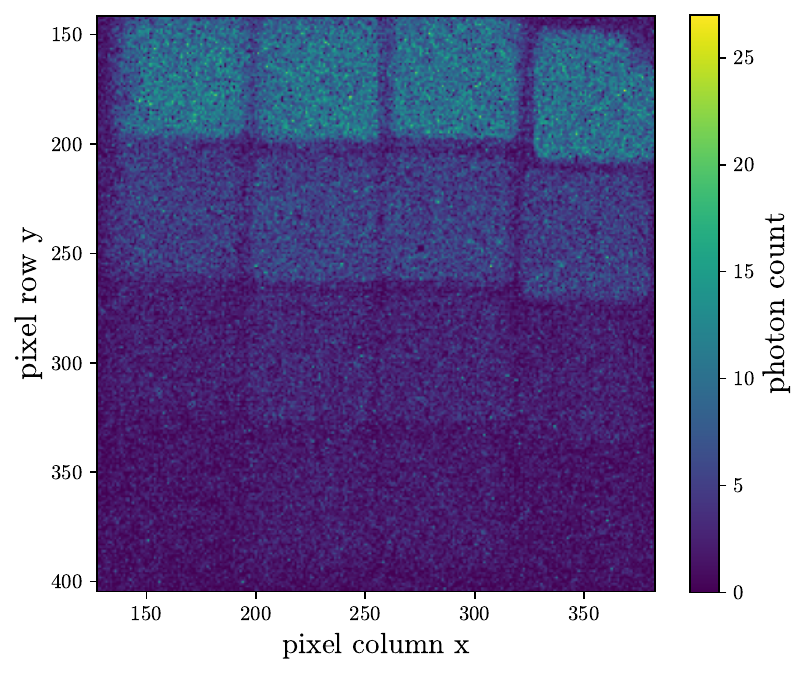}
            \caption*{Anomaly candidate frames summed.}
            \label{fig:both_ano2}
        \end{subfigure}
        \hfill
        \begin{subfigure}{0.45\textwidth}
            \centering
            \includegraphics[width=1.\linewidth]{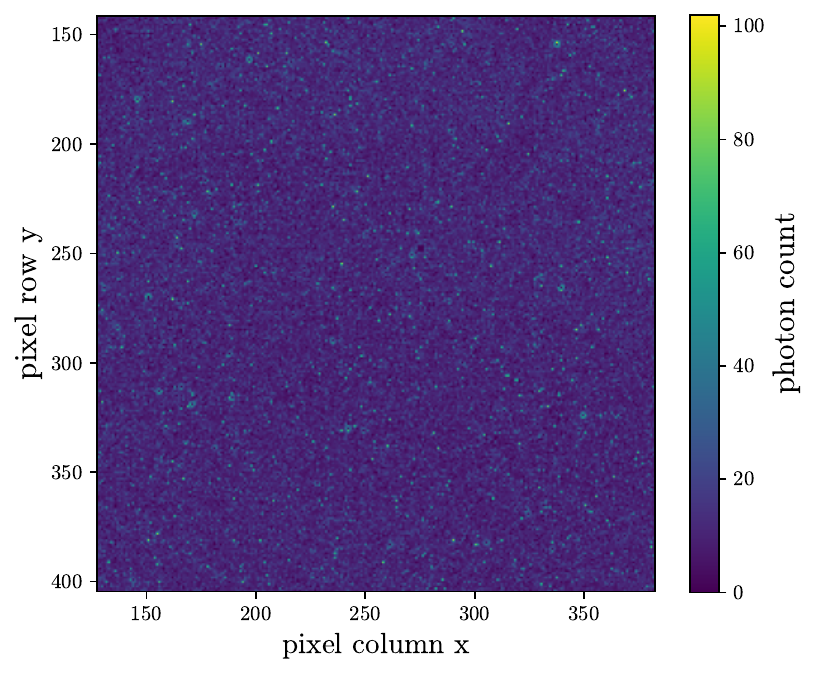}
            \caption*{Non-anomaly candidate frames summed.}
            \label{fig:both_nor2}
        \end{subfigure}
        \caption*{(b) 98th Percentile threshold ($\mathrm{BG}$ test set).}
        \label{fig:both_row2}
    \end{subfigure}

    \vspace{1em} 

    \begin{subfigure}{\textwidth}
        \centering
        \begin{subfigure}{0.45\textwidth}
            \centering
            \includegraphics[width=1.\linewidth]{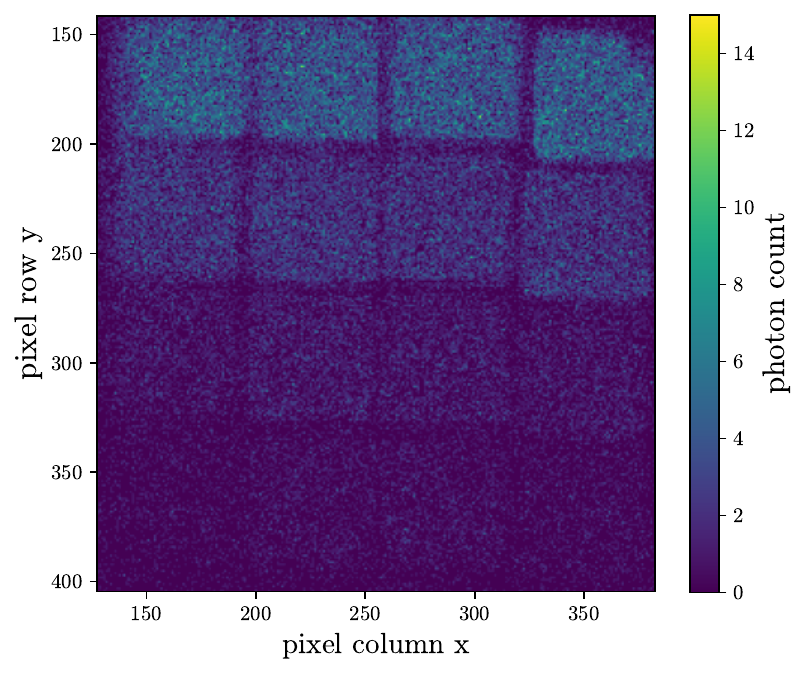}
            \caption*{Anomaly candidate frames summed.}
            \label{fig:both_ano3}
        \end{subfigure}
        \hfill
        \begin{subfigure}{0.45\textwidth}
            \centering
            \includegraphics[width=1.\linewidth]{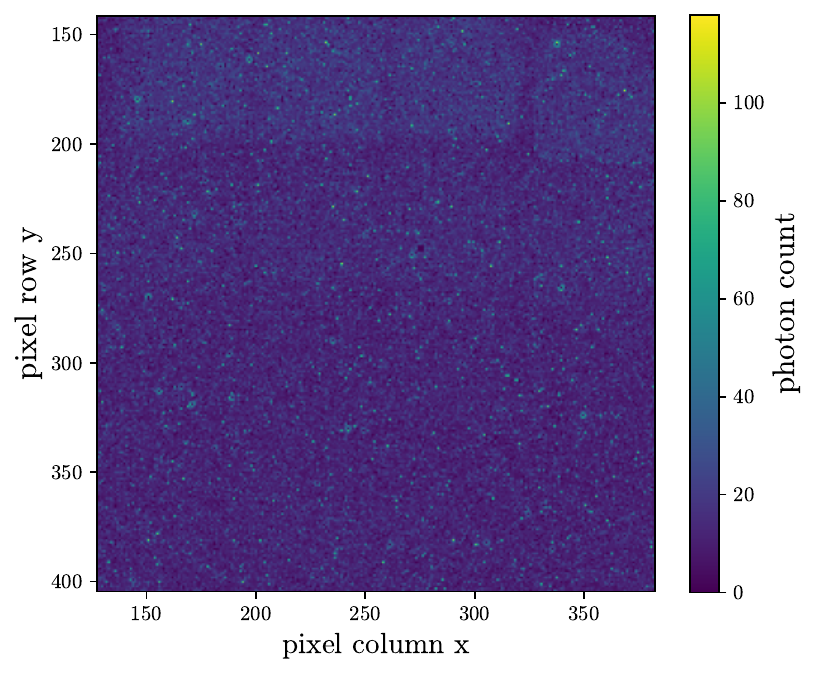}
            \caption*{Non-anomaly candidate frames summed.}
            \label{fig:both_nor3}
        \end{subfigure}
        \caption*{(c) Max threshold ($\mathrm{BG}$ test set).}
        \label{fig:both_row3}
    \end{subfigure}

    \caption{Summed frames within the fibre bundle region for the $^{90}\mathrm{Sr+BG}$ dataset, separated based on whether they are considered anomalies using both the BCE and the KLD losses.}
    \label{fig:both}
\end{figure}

Figure~\ref{fig:both} shows the total sum of frames selected and not selected as anomaly candidates based on different thresholds applied to either the BCE or KLD loss values, while Table~\ref{tab:both} presents the total number of frames selected in each case, categorised by the number of photon counts in the frame. The results suggest that combining both BCE and KLD performs similarly to using BCE alone. Future work could explore more sophisticated methods of combining these loss functions to leverage their strengths and improve anomaly detection efficiency.

\begin{table}[htbp]
    \centering
    \renewcommand{\arraystretch}{1.1} 
    \setlength{\tabcolsep}{8pt} 
    \resizebox{0.7\textwidth}{!}{ 
    \begin{tabular}{c|>{\columncolor{gray!20}}r|>{\columncolor{gray!20}}r|>{\columncolor{gray!20}}r|r|r|r}
        \toprule
        \multicolumn{7}{c}{\textbf{Divergence threshold ($^{90}\mathrm{Sr+BG}$ > $\mathrm{BG}$ test set)}} \\ \midrule
        \multirow{2}{*}{\textbf{Counts}} & \multicolumn{3}{c|}{\cellcolor{gray!30}\textbf{BG (test)}} & \multicolumn{3}{c}{\textbf{$^{90}\mathrm{Sr+BG}$}} \\ \cline{2-7} 
         & \textbf{Frames} & \textbf{Selected} & \textbf{\%} & \textbf{Frames} & \textbf{Selected} & \textbf{\%} \\ 
        \midrule
        4 & 23396 & 2253 & 9.63 & 153066 & 47297 & 30.90 \\
        \rowcolor{gray!10} 5 & 4830 & 3255 & 67.39 & 44146 & 39147 & 88.68 \\
        6 & 828 & 809 & 97.71 & 14397 & 14374 & 99.84 \\
        \rowcolor{gray!10} 7 & 109 & 109 & 100.00 & 7521 & 7521 & 100.00 \\
        8 & 14 & 14 & 100.00 & 5768 & 5768 & 100.00 \\
        \rowcolor{gray!10} 9 & 1 & 1 & 100.00 & 5159 & 5159 & 100.00 \\
        \bottomrule
        \toprule
        \multicolumn{7}{c}{\textbf{98th Percentile threshold ($\mathrm{BG}$ test set)}} \\ \midrule
        \multirow{2}{*}{\textbf{Counts}} & \multicolumn{3}{c|}{\cellcolor{gray!30}\textbf{BG (test)}} & \multicolumn{3}{c}{\textbf{$^{90}\mathrm{Sr+BG}$}} \\ \cline{2-7} 
         & \textbf{Frames} & \textbf{Selected} & \textbf{\%} & \textbf{Frames} & \textbf{Selected} & \textbf{\%} \\ 
        \midrule
        4 & 23396 & 104 & 0.44 & 153066 & 2944 & 1.92 \\
        \rowcolor{gray!10} 5 & 4830 & 367 & 7.60 & 44146 & 13630 & 30.87 \\
        6 & 828 & 424 & 51.21 & 14397 & 12457 & 86.52 \\
        \rowcolor{gray!10} 7 & 109 & 95 & 87.16 & 7521 & 7494 & 99.64 \\
        8 & 14 & 14 & 100.00 & 5768 & 5768 & 100.00 \\
        \rowcolor{gray!10} 9 & 1 & 1 & 100.00 & 5159 & 5159 & 100.00 \\
        \bottomrule
        \toprule
        \multicolumn{7}{c}{\textbf{Max threshold ($\mathrm{BG}$ test set)}} \\ \midrule
        \multirow{2}{*}{\textbf{Counts}} & \multicolumn{3}{c|}{\cellcolor{gray!30}\textbf{BG (test)}} & \multicolumn{3}{c}{\textbf{$^{90}\mathrm{Sr+BG}$}} \\ \cline{2-7} 
         & \textbf{Frames} & \textbf{Selected} & \textbf{\%} & \textbf{Frames} & \textbf{Selected} & \textbf{\%} \\ 
        \midrule
        4 & 23396 & 0 & 0.00 & 153066 & 4 & 0.00 \\
        \rowcolor{gray!10} 5 & 4830 & 0 & 0.00 & 44146 & 8 & 0.02 \\
        6 & 828 & 0 & 0.00 & 14397 & 177 & 1.23 \\
        \rowcolor{gray!10} 7 & 109 & 0 & 0.00 & 7521 & 1365 & 18.15 \\
        8 & 14 & 0 & 0.00 & 5768 & 3620 & 62.76 \\
        \rowcolor{gray!10} 9 & 1 & 0 & 0.00 & 5159 & 4833 & 93.68 \\
        \bottomrule
    \end{tabular}
    }
    
    \caption{Comparison of the fraction of selected frames using both the BCE and KLD losses for the $\mathrm{BG}$ testing set and the \textbf{$^{90}\mathrm{Sr+BG}$} sample for different numbers of counts per frame.}
    \label{tab:both}
\end{table}

\clearpage

\acknowledgments

We would like to thank the Edoardo Charbon and Claudio Bruschini and the Advanced Quantum Architecture Laboratory at EPFL for their help and fruitful discussions about the data acquisition for this project. Further, we would like to thank them for the provision of the SwissSPAD2 imaging sensor and the dataset used for testing the developed analysis method.

Part of this work was supported by the SNF grant PCEFP2\_203261, Switzerland.


\bibliographystyle{JHEP}
\bibliography{bibliography.bib}

\end{document}